\documentclass[10pt,conference]{IEEEtran}
\usepackage{cite}
\usepackage{amsmath,amssymb,amsfonts}
\usepackage{algorithmic}
\usepackage{graphicx}
\usepackage{textcomp}
\usepackage{xcolor}
\usepackage[hyphens]{url}
\usepackage{fancyhdr}
\usepackage{hyperref}
\usepackage{multirow}
\usepackage{comment}
\usepackage{color}
\usepackage[normalem]{ulem}
\usepackage[acronym,sort=standard,nonumberlist]{glossaries}
\glsdisablehyper 
\setglossarystyle{altlist} 
\hypersetup{
    colorlinks=true,
    linkcolor=blue
}
\newcommand{\hpcayear}{2026}

\makeglossaries
\newacronym{AD}{AD}{Autonomous Driving}
\newacronym{BOM}{BOM}{Bill Of Materials}
\newacronym{CCB}{CCB}{Central Control Block}
\newacronym{CNN}{CNN}{Convolutional Neural Network}
\newacronym{CRU}{CRU}{Clock and Reset Unit}
\newacronym{CSU}{CSU}{CPU Starter Unit}
\newacronym{CVU}{CVU}{Configurable Vector Unit}
\newacronym{DL}{DL}{Deep Learning}
\newacronym{DMA}{DMA}{Direct Memory Access}
\newacronym{DRB}{DRB}{Data Ring Bus}
\newacronym{DSA}{DSA}{Domain-Specific Architecture}
\newacronym{DTDU}{DTDU}{Data Transformation DMA Unit}
\newacronym{FSI}{FSI}{Functional Safety Island}
\newacronym{GSDU}{GSDU}{Gather/Scatter DMA Unit}
\newacronym{HBSM}{HBSM}{High Bandwidth Shared Memory}
\newacronym{ICB}{ICB}{Instruction Chain Bus}
\newacronym{ISP}{ISP}{Image Signal Processor}
\newacronym{JIT}{JIT}{Just-In-Time}
\newacronym{LLM}{LLM}{Large Language Model}
\newacronym{MAC}{MAC}{Multiply-ACcumulate}
\newacronym{MoE}{MoE}{Mixture-of-Experts}
\newacronym{NN}{NN}{Neural Network}
\newacronym{NoC}{NoC}{Network on Chip}
\newacronym{NPU}{NPU}{Neural Processing Unit}
\newacronym{PMU}{PMU}{Power Management Unit}
\newacronym{SC}{SC}{Synchronization Counter}
\newacronym{SU}{SU}{Synchronization Unit}
\newacronym{TCO}{TCO}{Total Cost of Ownership}
\newacronym{TCU}{TCU}{Tensor Computing Unit}
\newacronym{TPB}{TPB}{Tensor Processing Block}
\newacronym{TWU}{TWU}{Tensor Walker Unit}
\newacronym{VCIX}{VCIX}{Vector Core Instruction eXtension}
\newacronym{VLA}{VLA}{Vision-Language-Action}
\newacronym{VPU}{VPU}{Video Processing Unit}

\author{
  \ifdefined\hpcacameraready
    \IEEEauthorblockN{\hpcaauthors{}}
      \IEEEauthorblockA{
        \hpcaaffiliation{Li Auto}
        \vspace{-1.5em}
      }
  \else
    \IEEEauthorblockN{\normalsize{ISCA \hpcayear{} Submission
      \textbf{\#\hpcasubmissionnumber{}}} \\
      \IEEEauthorblockA{
        Confidential Draft \\
        Do NOT Distribute!!
      }
    }
  \fi 
}

\fancypagestyle{camerareadyfirstpage}{%
  \fancyhead{}
  
  \fancyhead[C]{
    \ifdefined\aeopen
    \parbox[][12mm][t]{13.5cm}{\hpcayear{} IEEE International Symposium on High-Performance Computer Architecture (HPCA)}    
    \else
      \ifdefined\aereviewed
      \parbox[][12mm][t]{13.5cm}{\hpcayear{} IEEE International Symposium on High-Performance Computer Architecture (HPCA)}
      \else
      \ifdefined\aereproduced
      \parbox[][12mm][t]{13.5cm}{\hpcayear{} IEEE International Symposium on High-Performance Computer Architecture (HPCA)}
      \else
      \parbox[][0mm][t]{13.5cm}{\hpcayear{} IEEE International Symposium on High-Performance Computer Architecture (HPCA)}
    \fi 
    \fi 
    \fi 
    \ifdefined\aeopen 
      \includegraphics[width=12mm,height=12mm]{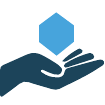}
    \fi 
    \ifdefined\aereviewed
      \includegraphics[width=12mm,height=12mm]{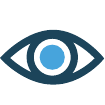}
    \fi 
    \ifdefined\aereproduced
      \includegraphics[width=12mm,height=12mm]{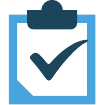}
    \fi
  }
  \fancyfoot[C]{}
}
\newcommand{\hpcasubmissionnumber}{82}
\title{M100: An Orchestrated Dataflow Architecture Powering General AI Computing}

\IEEEoverridecommandlockouts

\def\hpcacameraready{} 
\newcommand\hpcaauthors{
\parbox{\textwidth}{%
\centering
  Yan Xie$^{*}$\thanks{%
    $^{*}$Project lead. Other authors are listed in alphabetical order by first name. 
    Corresponding authors: Yan Xie (\texttt{xieyan@lixiang.com}) and Danyang Zhu (\texttt{zhudanyang@lixiang.com}).%
  },
  Changkui Mao, Changsong Wu, Chao Lu, Chao Suo, Cheng Qian, Chun Yang, 
  Danyang Zhu, Hengchang Xiong, Hongzhan Lu, Hongzhen Liu, Jiafu Liu, Jie Chen, Jie Dai, Junfeng Tang, Kai Liu, Kun Li, Lipeng Ge, Meng Sun, Min Luo, Peng Chen, Peng Wang, Shaodong Yang, Shibin Tang, Shibo Chen, Weikang Zhang, Xiao Ling, Xiaobo Du, Xin Wu, Yang Liu, Yi Jiang, Yihua Jin, Yin Huang, Yuli Zhang, Zhen Yuan, Zhiyuan Man, Zhongxiao Yao%
}
}
\newcommand\hpcaaffiliation[1]{#1}
\ifdefined\hpcacameraready 
\else
\fi

\ifdefined\eaopen

\fi

\begin{document}

\pagestyle{plain}
\setcounter{page}{1}
\maketitle
\renewcommand{\thefootnote}{}
\footnotetext{Accepted to appear at ISCA 2026 Industry Track.}
\renewcommand{\thefootnote}{\arabic{footnote}}

\begin{abstract}

 As deep learning-based AI technologies gain momentum across nearly all aspects of life, the demand for a general-purpose AI computing architecture continues to grow. While current GPGPU-based architectures offer exceptional versatility to diverse AI workloads, they often fall short in efficiency and cost-effectiveness. Conversely, various Domain-Specific Architectures (DSAs) excel at particular AI tasks but struggle to extend their capabilities across a broader range of applications, let alone adapt to the rapidly evolving landscape of AI algorithms.
 
 M100 is Li Auto's response to this challenge: a performant and cost-effective architecture designed to meet the AI inference demands of Autonomous Driving (AD), Large Language Models (LLMs), and intelligent human interactions. These domains are crucial to build today's most competitive automobile platforms. M100 employs a dataflow parallel architecture, where compiler-architecture co-design orchestrates not only computation but, more critically, data movement across time and space. Leveraging the inherent efficiency of dataflow computing, our integrated hardware-software approach improves overall system performance while significantly reducing hardware complexity and cost. In line with the dataflow paradigm, M100 largely eliminates caching. Tensor computations are driven by compiler- and runtime-managed data streams flowing between computing elements and on/off-chip memories, resulting in greater efficiency and scalability compared to conventional cache-based systems. Another key design principle was selecting the right operational granularity for scheduling, issuing, and execution—across the compiler, firmware, and hardware. Recognizing commonalities in AI workloads, we chose the tensor, large or small, as the fundamental data element in the M100 architecture. M100 has demonstrated general AI computing capability across diverse inference applications, including UniAD (for AD) and LLaMA (for LLMs). Benchmark results show that M100 outperforms GPGPU architectures in AD applications, with higher hardware utilization. We believe that M100 represents a promising direction for the future convergence of general AI computing architectures.
 
\end{abstract}

\section{Introduction}

Autonomous driving (AD) technology has been at the forefront of the AI technology evolution for quite some time. The cutting-edge vision-language-action (VLA) models\cite{black2024pi_0,brohan2022rt,zitkovich2023rt,wu2023unleashing,cheang2024gr} encompass many aspects of autonomous tasks, such as visual perception, environment comprehension, and action planning. The broad variety of AI inference tasks demands a versatile software and hardware solution that is not only performant but also adaptive to many forms of deep learning inference algorithms. Furthermore, the application environment inside an automobile, most likely electric, also necessitates accelerator architectures with small physical and power footprints. Li Auto has long recognized that an in-house developed AD accelerator chip is crucial in delivering automobile products that are competitive with both AD capabilities and Bill of Materials (BOM) cost.

Just like many other automobile manufacturers, Li Auto started developing its AD system based on off-the-shelf GPGPU platforms\cite{karumbunathan2022nvidia,nvidia_jetson_thor}. Even though these GPGPU platforms have been able to support earlier generations of Li Auto's AD system development and deployment with their advantages in general programmability and mature software  ecosystem, their limitations also emerge gradually in areas like peak performance, efficiency, customization and cost of ownership.  Major automobile manufacturers have chosen to develop in-house AD inference chips\cite{fsdhw3, fsdhw4, fsdhw5 } that are vertically integrated with their AD models and software stack. In order to reach the ultimate goal of providing customers with superb AD experience while keeping the BOM cost down, Li Auto also embarked on the journey of developing such an AI inference accelerator chip with an innovative architecture that meets all the performance and cost metrics. Additionally, this architecture should possess future-proofing characteristics that make it adaptable to the ever fast evolving AD models and algorithms. 

The fruit of this effort is M100 SoC integrated with the M100 NPU, an orchestrated dataflow architecture that delivers powerful general AI computing capabilities, proven by the AD tasks. We chose dataflow architecture because the majority of Deep Learning (DL) inference algorithms involve tensor computation and manipulation tasks, whose data movement and transformation patterns are generally regular and predictable.  Dataflow architecture\cite{dataflowSupercomputers,dataflowComputingModel,Think2020,soft2022Abts,prabhakar2017plasticine,prabhakar2021sambanova,prabhakar2024sambanova,Lie2024cerebras,lie2023cerebras,gwennap2020tenstorrent,vasiljevic2024tenstorrent,talpes2023microarchitecture,rico2024amd,perryman2023evaluation,moreira2020neuronflow,jouppi2017datacenter,jouppi2023tpu,firoozshahian2023mtia, coburn2025meta,baumgarte2003pact,govindaraju2011dynamically,singh2000morphosys} can effectively parallelize these tasks with minimal synchronization overhead. With the help of dataflow compiler, M100 managed to avoid the design complexity and overhead associated with the traditional dataflow architectures by allowing the compiler to orchestrate the task execution at a higher granularity, hence we call M100 an ``orchestrated dataflow architecture''. The success of the M100 architecture also requires Li Auto team to make the right trade-offs while balancing the complexity of the software and hardware, selecting the granularity of the accelerated operations, and choosing the degree of deterministic and non-deterministic behavior of the hardware components. In our view, Li Auto's M100 architecture may have landed on the sweet spot in addressing the general AI inference computing challenges. 


The remainder of the paper presents the M100 architecture and its application outcomes. Section \ref{Motivation} outlines the motivation behind Li Auto’s development of an in-house AI inference chip. Section \ref{ArchitectureIntro} explains the design principles guiding the M100 architecture. Section \ref{ArchitectureDetails} provides a detailed description of the M100 NPU. Section \ref{SoftwareStack} briefly introduces the compiler and software stack. Section \ref{EvaluationResults} presents evaluation methods and real-world results. Finally, Section \ref{Conclusion} summarizes Li Auto’s efforts and discusses future directions for the M100 project.

\vspace{-1.5em}
\section{Motivation}
\label{Motivation}

Deep learning–based AD systems rely on neural networks for perception, prediction, and planning using camera images and LiDAR data. These tasks are highly compute-intensive and must be executed with low latency to ensure safe operation at high speeds. The GPGPU-based platforms such as NVIDIA’s Orin\cite{karumbunathan2022nvidia} and Thor\cite{nvidia_jetson_thor} are built upon SIMT architecture\cite{nickolls2010gpu} augmented with the tensor cores\cite{jia2018dissecting}. While they are widely used for their versatility and strong parallel processing capabilities, they come with trade-offs. These off-the-shelf solutions are not tailored to specific AD software stacks, often include unused features, and carry a high total cost of ownership (TCO). Their cache-based memory hierarchy also introduces optimization challenges and unpredictability. In response, some companies have turned to Domain-Specific Architectures (DSAs), such as Tesla’s FSD chip\cite{fsdhw3,fsdhw4,fsdhw5}, which hardwire neural network operations into fixed pipelines. While highly efficient for targeted tasks, DSAs often struggle to keep pace with rapidly evolving AI algorithms—especially with the rise of end-to-end VLA models—leading to shorter lifecycles and higher reengineering costs.

Recognizing the need for a middle ground, Li Auto set out to design an NPU architecture that balances efficiency with flexibility. The result is M100—a scalable, dataflow-driven architecture built to support a broad spectrum of edge AI inference tasks. Its modular design and hierarchical interconnect enable high hardware utilization and adaptability across vehicle generations, helping to amortize development costs while maintaining performance leadership in the face of fast-changing AI demands.

\section{Orchestrated Dataflow Architecture}
\label{ArchitectureIntro}

\subsection{Design Philosophy}
Departing from traditional instruction-sequenced execution models of CPUs and GPUs, M100 NPU adopted the data-driven parallel execution model\cite{suettlerlein2013implementation,theobald1999earth}. Instead of executing predefined instruction streams, M100 NPU distributes tensor operation instructions to a large number of execution units, among which data flows and triggers operations of those instructions. To further increase M100 NPU's capacity, homogeneous computing nodes, each of which is capable of running the complete set of tensor instructions, are interconnected with a scalable network optimized for inter-node data movement and synchronization. Within each node, data and synchronization paths among various execution units can also be flexibly constructed to support intra-node dataflow execution. With its modular and scalable design, the M100 NPU architecture strives to provide an elastic hardware abstraction layer on which the compiler can map an AI inference task and orchestrate its execution with optimal performance. The following sections discuss how design decisions were made for various aspects of the M100 NPU architecture.

\subsubsection{Computing Elements}
\begin{figure}[ht]
	\centering
    \vspace{-1em}
	\centerline{\includegraphics[scale=0.35]{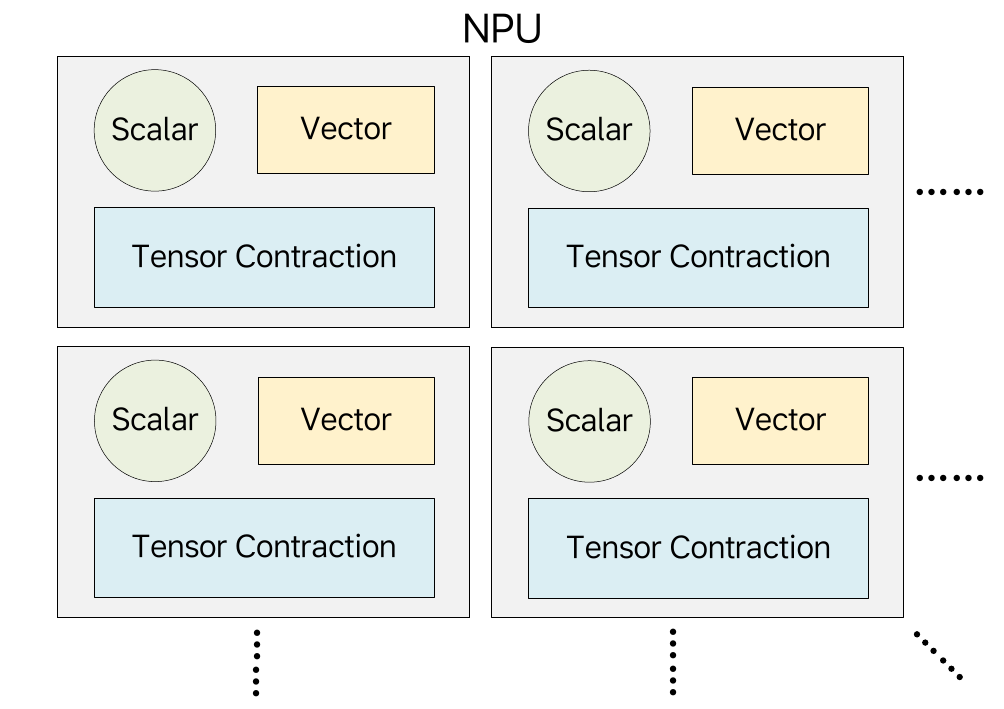}}
	\caption{Computing blocks in M100, each composed of three computing elements.}
	\label{NPU computing elements}
\end{figure}

The M100 NPU is built to accelerate a wide range of deep learning inference tasks used in autonomous driving, many of which rely heavily on tensor contraction operations such as convolution and matrix multiplication—requiring compute-dense functional units for high throughput. In addition, vector operations, though less compute-intensive, involve a wide variety of operations, demanding a balance between flexibility and performance. Scalar computations are also common and require general-purpose CPU cores. As illustrated in Fig. \ref{NPU computing elements}, M100 integrates tensor, vector, and scalar processing units into unified computing blocks with shared local memory and synchronization. The architecture scales by instantiating multiple such blocks connected via on-chip communication networks, with software orchestrating coarse-grained instruction dispatch across them.

\subsubsection{Memory Hierarchy}

Parallelism remains the primary strategy for accelerating AI inference workloads, but system performance heavily depends on how data is shared across parallel execution units. Cache-coherent memory systems simplify programming by abstracting a large shared memory space, leveraging temporal and spatial locality when possible. However, these systems struggle to scale in massively parallel environments and often hinder streaming performance—a key aspect of AI inference. To address this, M100 adopts a modernized dataflow computing model.

\begin{figure}[ht]
	\centering
	\centerline{\includegraphics[scale=0.25]{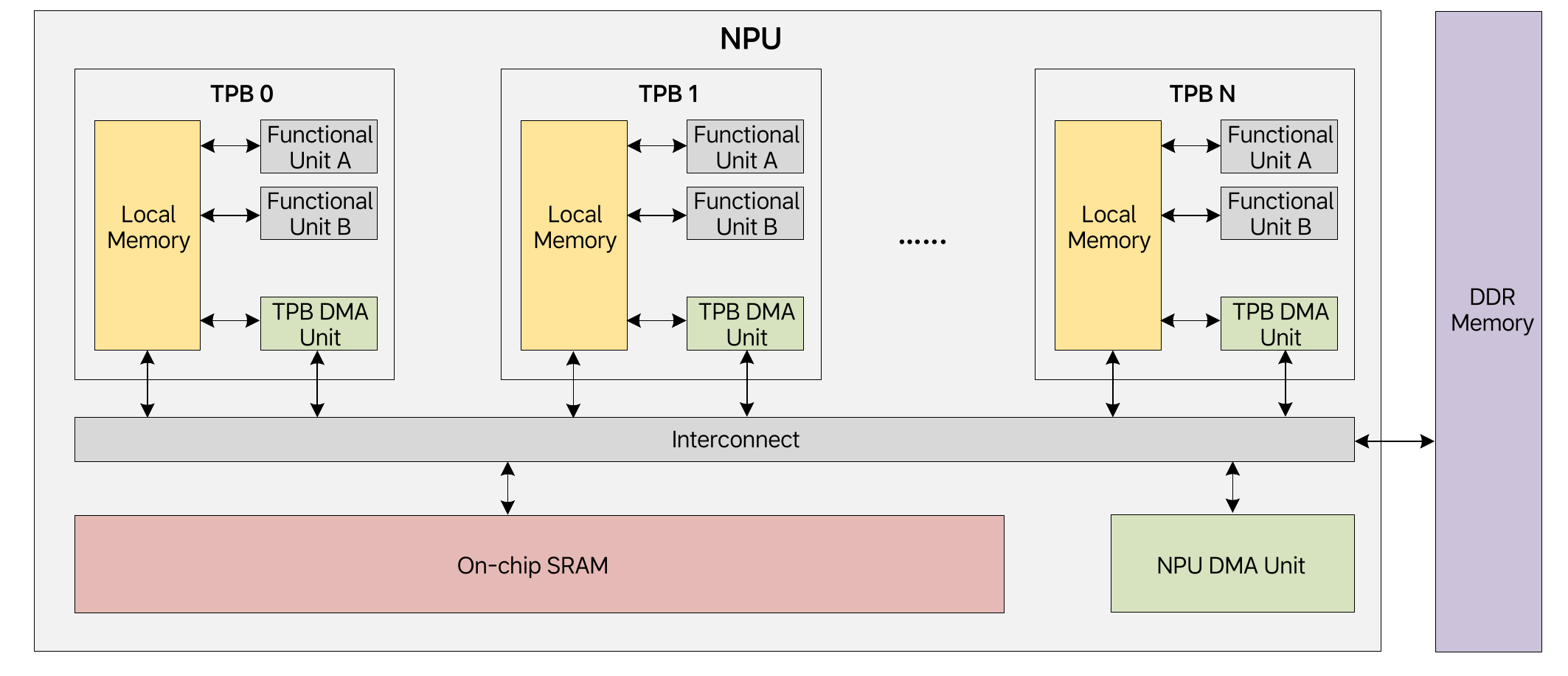}}
	\caption{ Architecture of the M100 NPU memory system without multi-level caches.}
	\label{NPU design}
\end{figure}

As shown in Fig. \ref{NPU design}, the M100 NPU largely avoids multi-level caches. Each Tensor Processing Block (TPB) includes high-bandwidth local memory, enabling functional units to stream data in and out in parallel during compute and manipulation tasks. Data transfers between TPB memories and the NPU’s shared SRAM are explicitly controlled via programmable DMA units. Additional DMA engines manage transfers between SRAM and DDR memory. This software-managed data movement—combined with efficient dataflow synchronization and sufficient buffering—allows computation and data transfer to overlap, maximizing throughput.

\subsubsection{Operation Granularity}

Since most AI inference workloads involve tensor-based computation and data transfer, it is natural to define accelerator instructions at the tensor level. This enables a streaming architecture where operands and results flow directly to and from memory, eliminating the need for register files and explicit load/store instructions. Memory latency is amortized over large tensors, and pipelined execution maximizes throughput. While some irregular operations still require fine-grained computation on conventional CPU cores, these tasks are generally not on the critical path. Therefore, M100 dedicates most hardware resources to regular, tensor-granularity computation, supplemented by lightweight CPU cores to address fine-grained general-purpose computing needs.

\subsubsection{ Data Flow Synchronization}

\begin{figure}[ht]
	\centering
	\centerline{\includegraphics[scale=0.35]{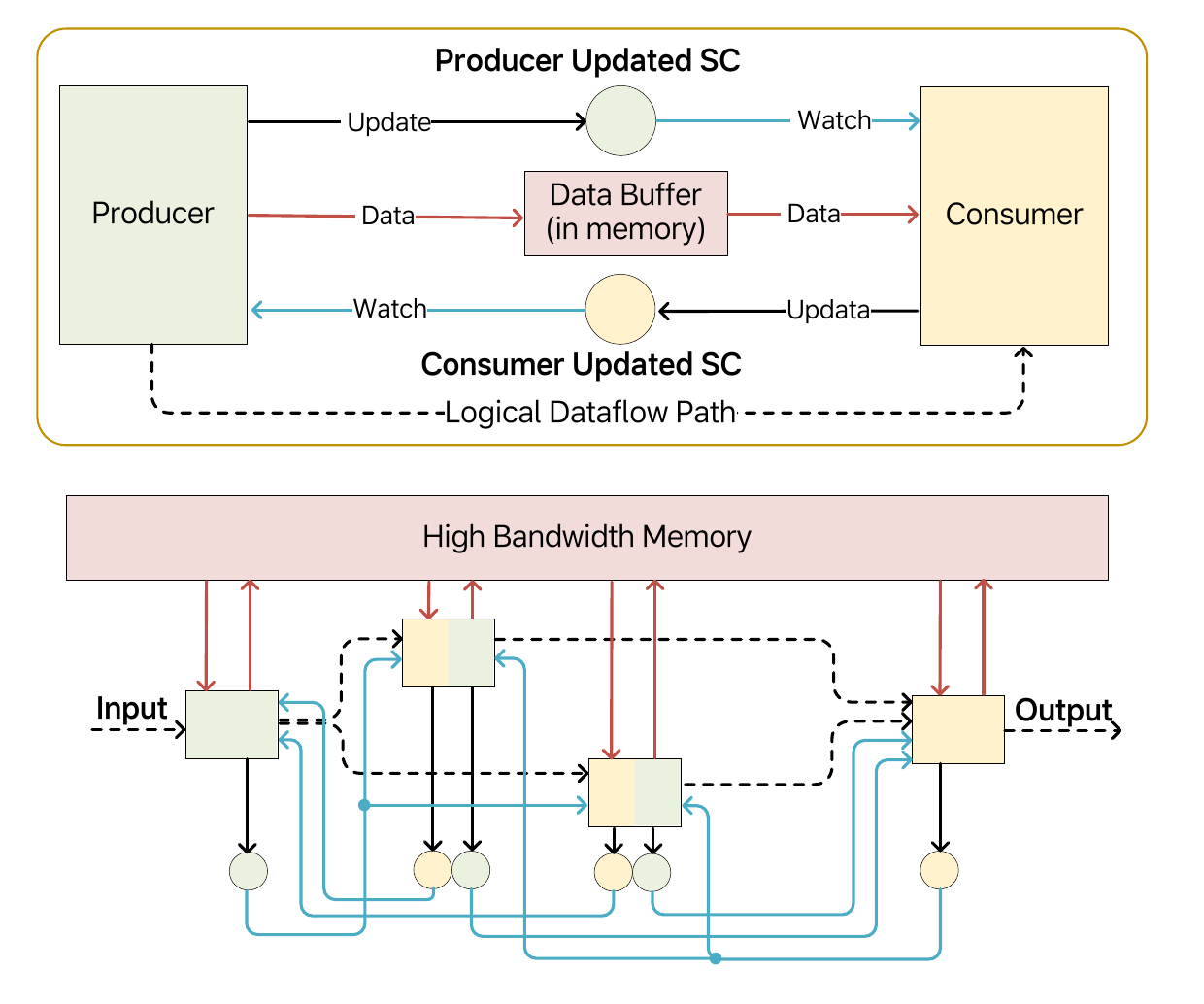}}
    \vspace{-1.5em}
	\caption{Two-way producer/consumer synchronization scheme for concurrent processing engines.}
	\label{Dataflow Computing Graph}
\end{figure}

Another key aspect of the M100 NPU design is its efficient synchronization mechanism. As a massively parallel system, M100 coordinates many concurrent processing engines through a producer–consumer synchronization model, illustrated in Fig. \ref{Dataflow Computing Graph}.\label{Section III-C-4} {The upper half of the diagram illustrates the producer-consumer synchronization between two agents. The red arrowed lines represent the memory read and write operations. The black arrowed lines represent updates to the Synchronization Counters (SCs), and the blue arrowed lines represent the monitoring behavior on the SCs. The dotted arrowed line represents the logical direction of the data movement from the producer to the consumer.} The producer writes data to a preallocated buffer and signals availability by updating an SC. The consumer monitors this SC and begins processing once the expected data is available. Conversely, the consumer updates a separate SC to inform the producer when buffer space is freed, enabling continued data flow. These SC operations are handled by dedicated hardware, ensuring minimal synchronization overhead. Synchronization granularity is software-controlled, allowing flexibility in how frequently producers and consumers coordinate during tensor operations. The lower half of the diagram shows how this SC based synchronization mechanism can be extended to a group of agents, some of whom may act as both producer and consumer—effectively making the M100 a dataflow-parallel computing system.

M100 NPU also supports other synchronization patterns such as barrier, broadcast, and reduction. These synchronization mechanisms are highly efficient and easily programmable, which works not only within one NPU but also across multiple NPUs in multi-chip configuration.

\subsubsection{Instruction Dispatch}

The M100 NPU uses a centralized instruction dispatcher, leveraging the Instruction Chain Bus to broadcast tensor operation instructions efficiently across multiple processing elements. To simplify hardware design, instructions for each processing element must execute in dispatch order, while instructions across different elements may complete out of order. It is the software's responsibility to manage synchronization when dependencies exist. Unlike traditional dataflow architectures, this design shifts some complexity to the compiler and runtime, which can take advantage of the regularity in AI inference workloads to plan and schedule execution. This “Orchestrated Dataflow Architecture” strikes a practical balance between hardware simplicity and software control, while retaining the efficiency of dataflow parallelism.

\subsubsection{Summary}

In summary, the M100 NPU integrates tensor/vector compute engines, DMA units, and lightweight CPU cores. Most computation is performed at the tensor level in a streaming fashion, with data flowing directly to and from memory. General-purpose tasks are handled by lightweight CPUs, potentially with vector extensions. The compiler orchestrates dataflow execution by dispatching compute and data movement instructions and managing synchronization across processing elements. Architectural details are discussed in the following section.

\subsection{M100 Overview}

\subsubsection{M100 SoC}

\begin{figure}[ht]
	\centering
	\centerline{\includegraphics[scale=0.23]{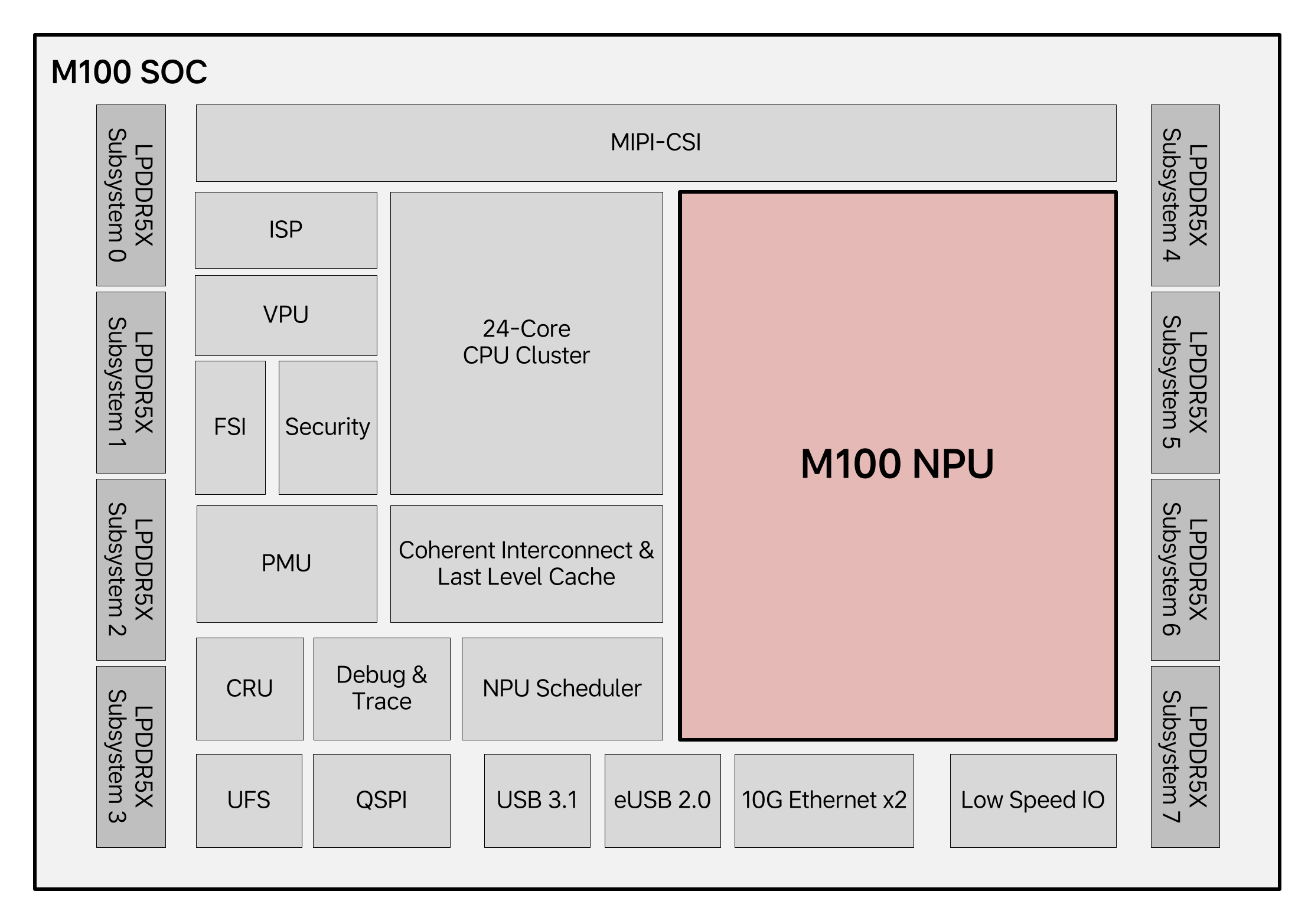}}
	\caption{The high level block diagram of M100 SoC.}
	\label{M100 SoC}
\end{figure}

M100 is an SoC designed to support Li Auto’s AD software stack. Like other AD chips, it includes application CPUs, multimedia IPs, a security island, and standard I/O interfaces. Its key differentiator is the indigenously developed Neural Processing Unit (NPU), built by Li Auto to accelerate AI inference. Figure \ref{M100 SoC} shows the high-level block diagram.
	
Figure \ref{M100 SoC} highlights the major functional blocks of the M100 SoC. It features 8 LPDDR5X subsystems, delivering 64 GB of memory and 273 GB/s peak bandwidth. The MIPI-CSI system supports input from up to 11 cameras, with an Image Signal Processor (ISP) subsystem that processes raw images for the NPU’s perception models. A video processing unit (VPU) handles video encoding/decoding, while the Functional Safety Island (FSI) and Security Engine ensure functional safety (FuSa) compliance and secure content handling. The Power Management Unit (PMU) and Clock and Reset Unit (CRU) coordinate power-up sequencing and clock/reset distribution. A dedicated NPU Scheduler dispatches inference tasks and collects results. Debug \& Trace modules support intrusive and non-intrusive debugging across CPU and NPU subsystems. The SoC also includes UFS/QSPI controllers for external storage, USB/Ethernet for high-speed I/O, and various low-speed interfaces. The CPU cluster integrates 24 ARM Cortex-A78AE cores with coherent CMN interconnect and a shared last-level cache.

\subsubsection{M100 NPU}

\begin{figure}[ht]
	\centering
	\centerline{\includegraphics[scale=0.19]{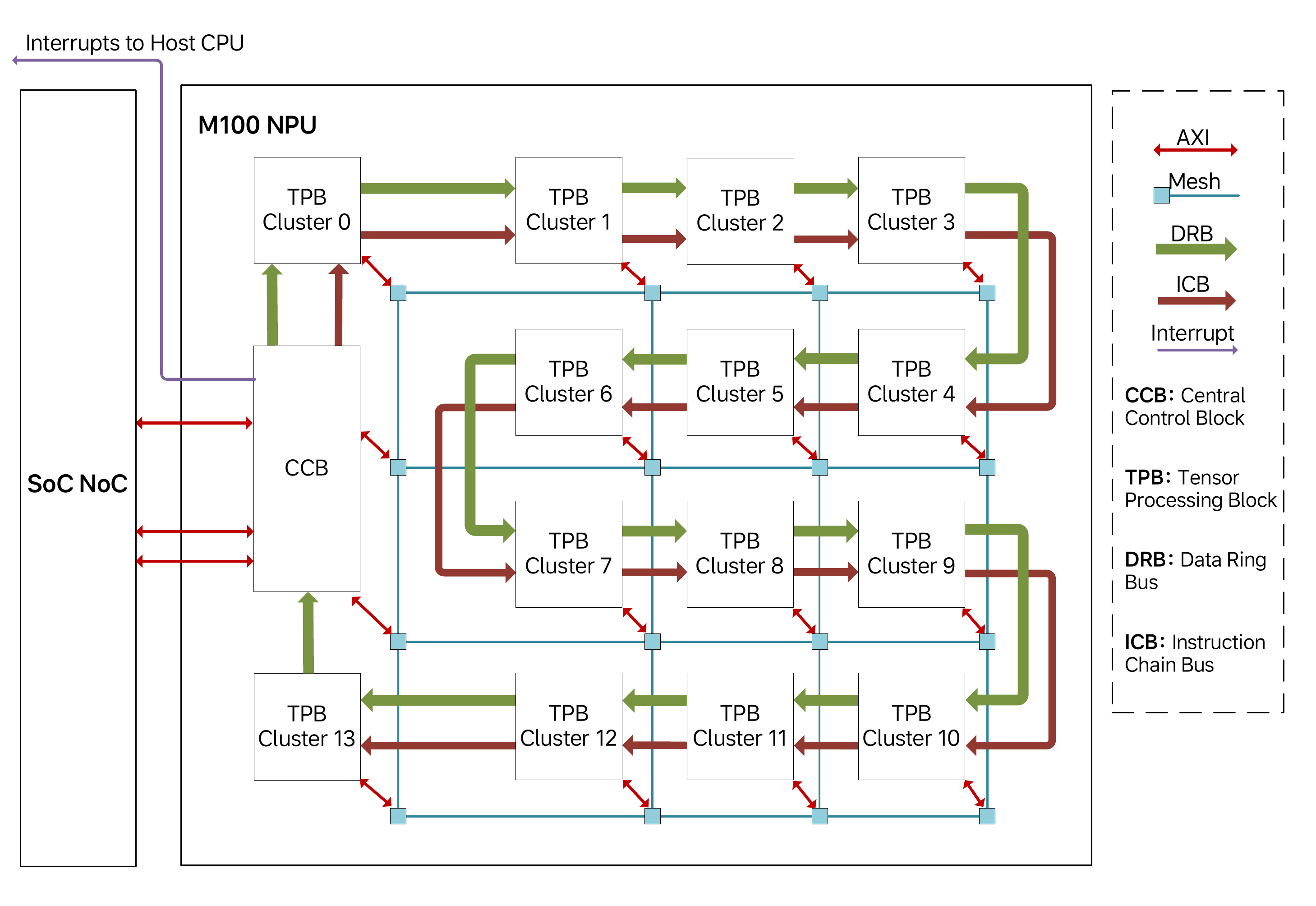}}
	\caption{The high level architecture of M100 NPU.}
	\label{M100 NPU Top}
    \vspace{-1.5em}
\end{figure}

The M100 NPU, central to this paper, is the most important subsystem in the SoC. It occupies a significant portion of the die and serves as the primary engine for AI inference. Its innovative dataflow architecture sets M100 apart from other AI accelerators. Figure \ref{M100 NPU Top} shows the NPU’s high-level architecture.

The NPU connects to the rest of the SoC through three main interfaces. First, two high-bandwidth AXI master interfaces (128 GB/s each) enable access to DDR memory and other SoC resources via the NoC system, which supports enough outstanding transactions to sustain peak memory throughput. Second, the NPU can generate interrupts to notify the scheduler CPU of events, such as task completion. Third, the scheduler CPU communicates with the NPU through a lower-bandwidth AXI slave interface to issue commands, check status, and access internal resources.

Internally, the NPU consists of one Central Control Block (CCB) and 14 Tensor Processing Block (TPB) clusters, each containing 4 TPBs. To support the data movement needs of AI inference, the CCB and TPBs are connected by two interconnects: a 2D Mesh Bus and a Data Ring Bus (DRB). The Mesh Bus offers scalable, high-bandwidth point-to-point communication\label{Section III-A-2} among TPB clusters, Central Control Block, CPUs, DMAs, and Block SRAMs. It provides up to 256 GB/s per node pair—scaling well under low-congestion conditions. The DRB, on the other hand, provides a deterministic, high-efficiency broadcast path with up to 256 GB/s aggregated bandwidth, making it ideal for multicasting data across TPBs. The software dynamically selects between the Mesh and DRB interconnects based on communication needs.



The Instruction Chain Bus (ICB) links the CCB to TPB clusters in a daisy-chain fashion. RISC-V cores in the CCB dispatch instructions to individual or multiple TPBs via the ICB. These TPB instructions define tensor operations and include rich metadata such as tensor shape and communication requirements. Though each instruction can be thousands of bits long—taking hundreds of cycles to transmit at 64 bits/cycle—their execution time spans tens of thousands of cycles, ensuring instruction dispatch is not a bottleneck.

The following sections delve into the architectural details of the M100 NPU’s building blocks, highlighting how its dataflow execution model and carefully chosen programming granularity enable both high performance and flexibility.

\section{NPU Architecture Details}
\label{ArchitectureDetails}

\subsection {Central Control Block} 

\begin{figure}[ht]
	\centering
	\centerline{\includegraphics[scale=0.22]{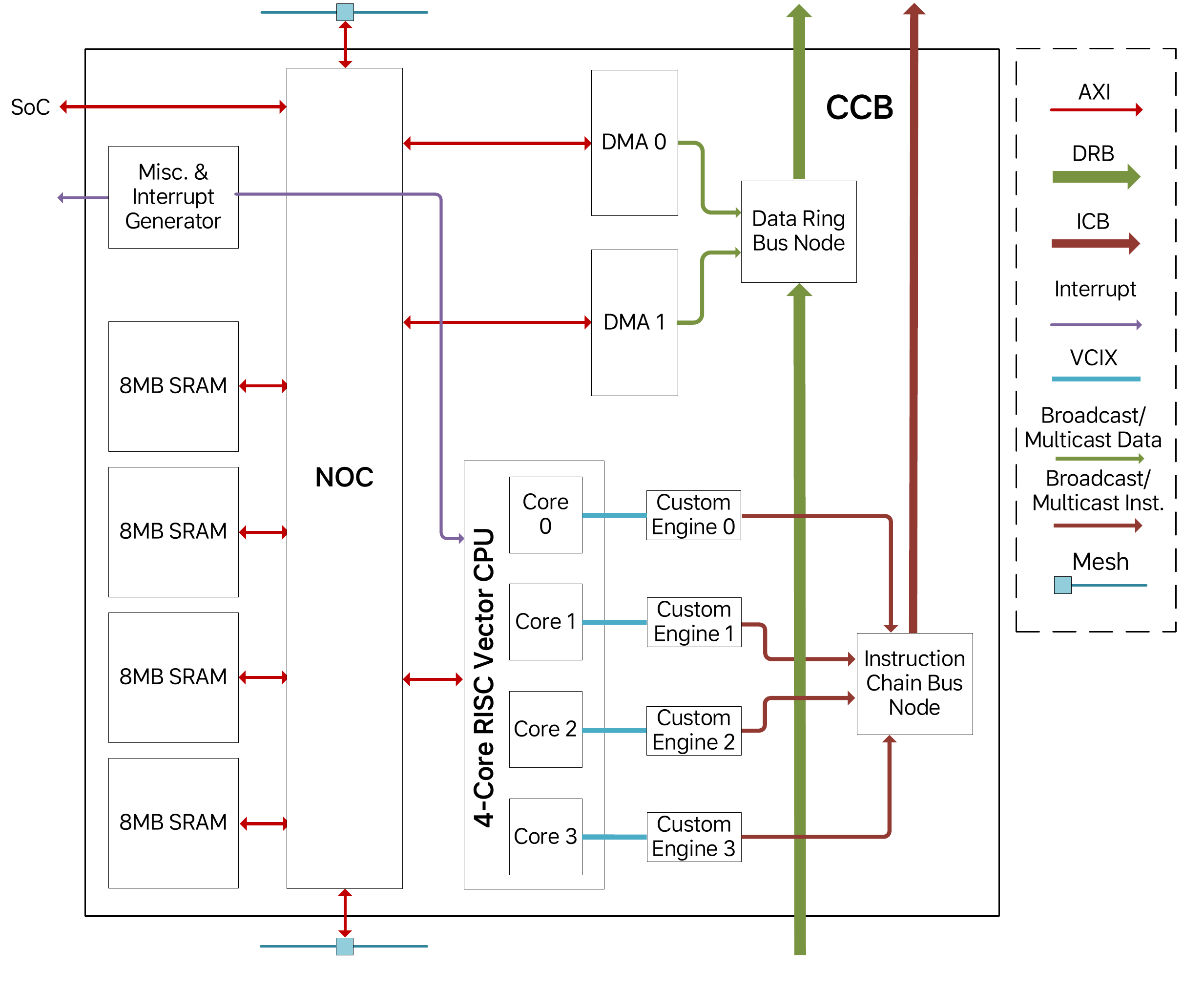}}
	\caption{Architecture of the CCB.}
	\label{CCB}
    \vspace{-1em}
\end{figure}

The Central Control Block (CCB), shown in Fig. \ref{CCB}, serves as the control center of the NPU. Its firmware runs on a 4-core SiFive X280 RISC-V CPU, each paired with a custom vector engine that dispatches TPB instructions via the ICB. These engines parse and forward large, complex TPB instructions—often thousands of bits long—to define tensor operations such as matrix multiplication or element-wise addition. Instructions include operand access, computation method, and result handling. With four CPU–engine pairs, the CCB supports up to four concurrent inference tasks. TPB instructions can also be broadcast to multiple TPBs using a destination mask, and given their long execution time, ICB bandwidth is typically sufficient for sustained throughput. The CCB includes 32 MB of on-chip SRAM, split into four 8 MB banks with 4 KB interleaving to enable high-bandwidth parallel access. Two DMA engines manage data transfers between the DDR and the CCB SRAM, and can also broadcast weights to TPBs directly via the DRB, which supports up to 256 GB/s, matching DDR read bandwidth.
Additional CCB functions include barrier synchronization and interrupt generation. Barrier operations ensure groups of TPBs complete their current instructions before continuing, useful for infrequent global sync points. Interrupts can be triggered to CCB or host CPUs via control registers. All components are interconnected via Arteris FlexNoC.

\subsection {Tensor Processing Block Cluster} 

\begin{figure}[ht]
	\centering
	\centerline{\includegraphics[scale=0.21]{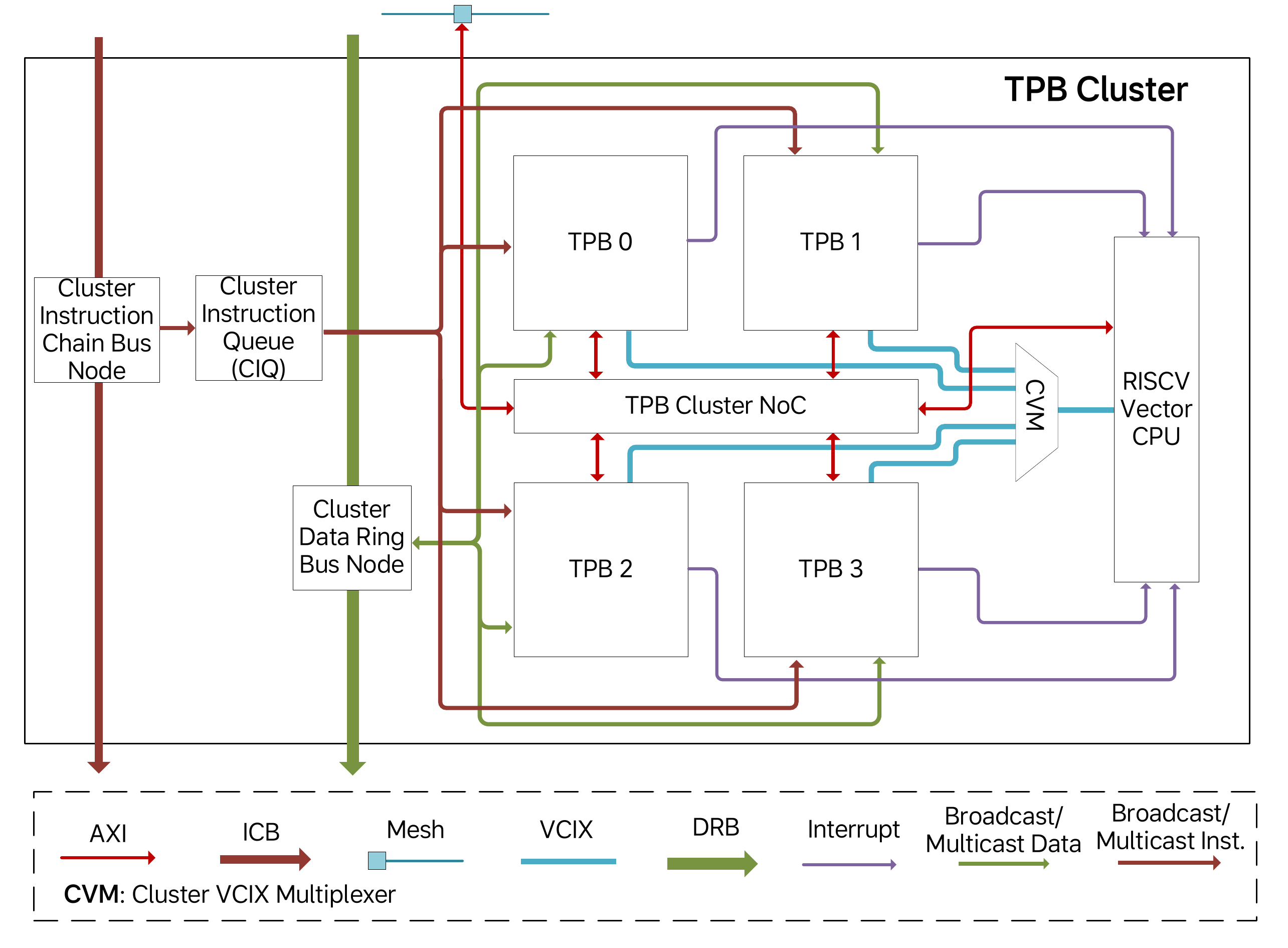}}
	\caption{Architecture of the TPB Cluster.}
	\label{TPB Cluster}
    \vspace{-1.5em}
\end{figure}


Figure \ref{TPB Cluster} illustrates the structure of a TPB cluster. There are two main reasons for introducing a cluster-level hierarchy. First, four TPBs can share common resources—such as the instruction buffer, ICB and DRB nodes, and a RISC-V CPU—allowing more silicon area to be allocated to tensor processing, thus improving compute density. Second, close proximity among the four TPBs enables low-latency, high-bandwidth communication, making it ideal for tasks that span a small number of TPBs—a common case in our AD inference workloads. For larger tasks, multiple clusters can collaborate via the Mesh Bus, though programmers should be mindful of the relatively lower communication efficiency between TPBs in different clusters and apply optimizations accordingly.


The shared RISC-V Vector CPU (SiFive X280) contributes general-purpose compute. TPB instructions can trigger CPU-based tasks via interrupts. The CPU retrieves task parameters, executes preloaded service routines, and marks the TPB instruction complete upon finishing. Up to four TPBs can request service concurrently, and the CPU arbitrates and handles requests sequentially. This mechanism allows CPU-based operations to follow the same instruction semantics as tensor operations, simplifying compilation, scheduling, and dispatch.


Each cluster includes a TPB instruction queue that downloads instructions from the ICB and stores them in a large buffer. Instructions are dispatched to TPB functional units as they become ready, without requiring global execution order—only preserving order within the same functional unit of a single TPB. This reflects the core of our Orchestrated Dataflow Architecture, where the compiler emits a loosely sorted instruction stream, and runtime execution is driven by data readiness and synchronization conditions. The instruction queue ensures functional units remain busy as soon as inputs and syncs are satisfied.


Similar to the CCB, each cluster includes an internal NoC connecting the four TPBs and the CPU memory port. The cluster NoC links to the NPU-level Mesh Bus through master/slave ports for bidirectional data access. The ICB node handles TPB instruction delivery, while the DRB node manages broadcast data traffic to and from the cluster.

\subsection{Tensor Processing Block}\label{TPB section}

\begin{figure}[ht]
	\centering
	\centerline{\includegraphics[scale=0.22]{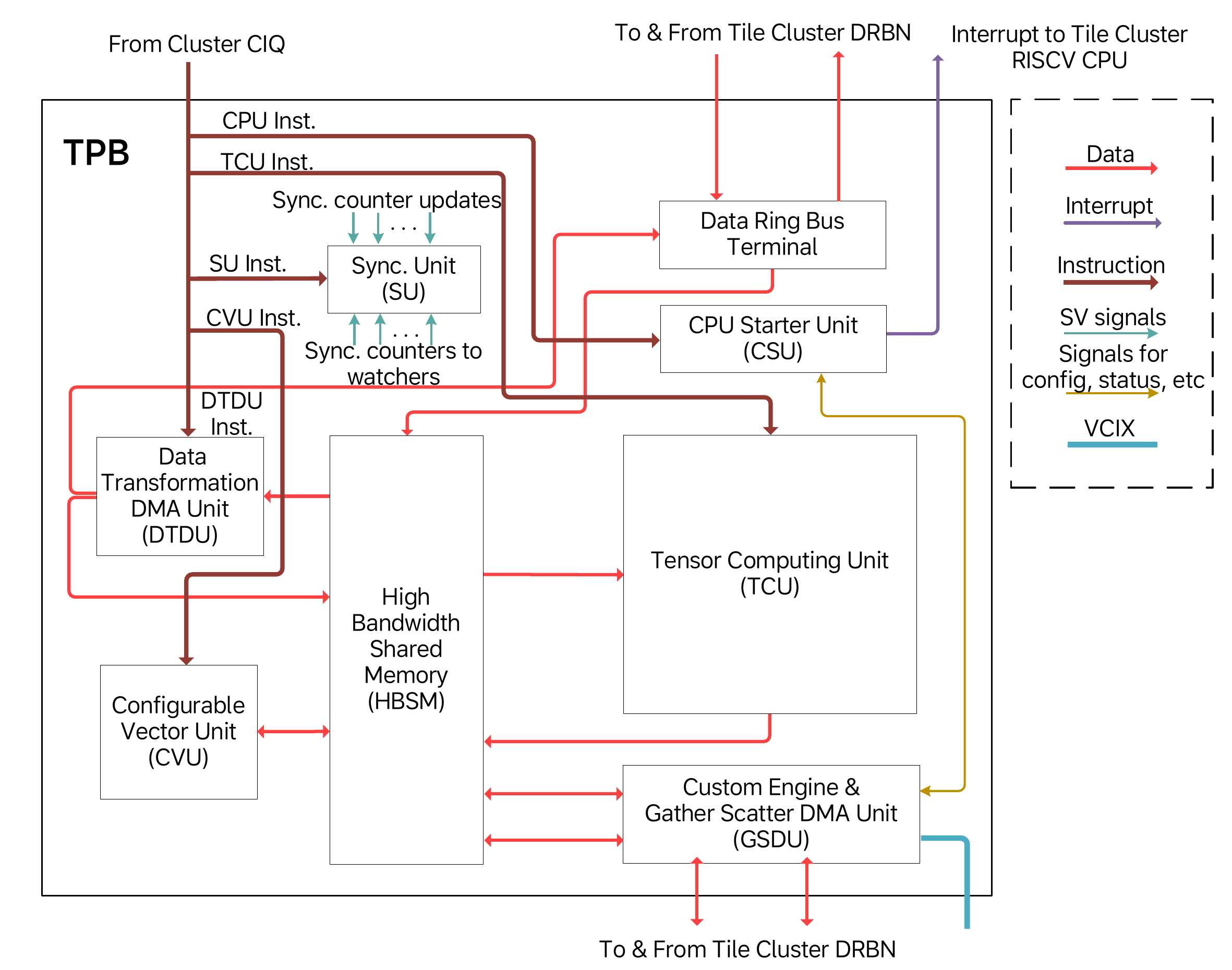}}
	\caption{Architecture of the TPB.}
	\label{fig:TPB}
    \vspace{-1.5em}
\end{figure}


The TPB is the core unit responsible for tensor computation and transformation. As shown in Fig. \ref{fig:TPB}, it consists of several specialized functional units, each optimized for a specific type of tensor operation. Below is a brief overview of the major functional units within the TPB:

\begin{itemize}

    \item  The High Bandwidth Shared Memory (HBSM) serves both as a 2 MB data store and a flexible communication hub for TPB functional units. Producers and consumers exchange data via predefined address ranges, synchronized through counters—eliminating the need for dedicated datapaths. To reduce SRAM port conflicts and maintain performance, HBSM uses a banked memory design.
    

    \item  The Tensor Computing Unit (TCU) handles the most compute-intensive operations, such as convolution and matrix multiplication, and includes a nonlinear pipeline for activation functions.


    \item The Configurable Vector Unit (CVU) consists of modular vector arithmetic units that can be reconfigured into custom pipelines. It efficiently handles basic vector operations and common AI tasks such as pooling, softmax, and layer normalization.


    \item The Data Transformation DMA Unit (DTDU) handles data movement within the TPB or broadcasts to other TPBs. It also supports tensor layout transformations, such as matrix transposition.
    

    \item The CPU Starter Unit (CSU) handles TPB instructions that request cluster CPU assistance. It saves the instruction parameters and triggers an interrupt. The CPU then accesses the requesting TPB’s data and devices via the Vector Coprocessor Instruction eXtension (VCIX) interface.


    \item The Custom Engine executes TPB operations on behalf of the CPU via the VCIX interface, including control register and memory access. It also includes a Gather/Scatter DMA Unit (GSDU), which the CPU can use to move data with non-contiguous address patterns.


    \item The Synchronization Unit (SU) manages synchronization counters that TPB functional units can update or monitor locally. It also supports remote synchronization via the DRB and NPU NoC.
    
\end{itemize}
The following sections provide a more detailed discussion of the TPB functional units.

\textbf{1) High-bandwidth Shared Memory Unit}

\begin{figure}[ht]
	\centering
	\centerline{\includegraphics[scale=0.23]{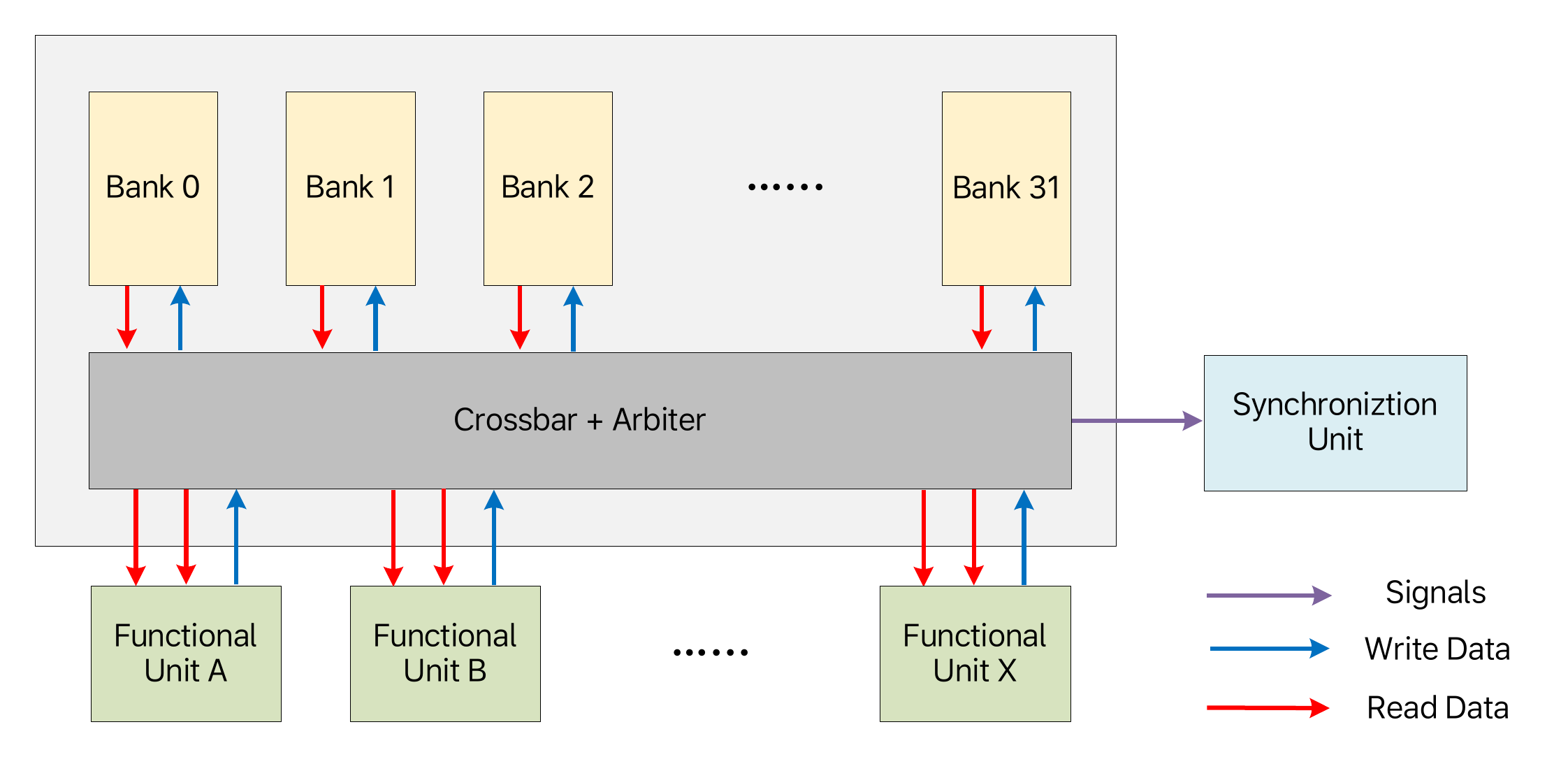}}
	\caption{Architecture of the HBSM.}
	\label{HSMU}
\end{figure}


The 2MB HBSM SRAM is uniformly shared across all TPB functional units. As shown in Fig. \ref{HSMU}, most units stream tensors in and out of HBSM—often in parallel—while executing tasks. Since outputs from one unit frequently serve as inputs to another, HBSM enables efficient producer–consumer communication without requiring dedicated datapaths. While the shared memory introduces latency ($\sim$ 20 cycles), the streaming nature of TPB operations minimizes its impact, provided high bandwidth is maintained—not just for individual units, but for concurrent access by multiple units, which is critical to sustaining TPB throughput.


HBSM achieves high bandwidth through a banked architecture, using 32 memory banks, each supporting 32 bytes per cycle. Address space is interleaved at 32-byte granularity, enabling simultaneous access across banks. Although more banks improve bandwidth by reducing collisions, they also increase routing congestion—especially in high-throughput designs. After extensive modeling and backend evaluation, a configuration of 32 banks and 8 requester ports was chosen as the optimal balance.


When multiple requesters target the same bank, HBSM uses round-robin arbitration and guarantees in-order access per requester. Synchronization actions—such as marking data as produced or consumed—are tied to memory accesses and triggered once arbitration is won. From that point, the access is considered globally visible, as no later request can overtake it. By unifying data movement and synchronization, HBSM serves as the central backbone of the M100’s dataflow architecture.

\begin{figure}[ht]
	\centering
	\centerline{\includegraphics[scale=0.33]{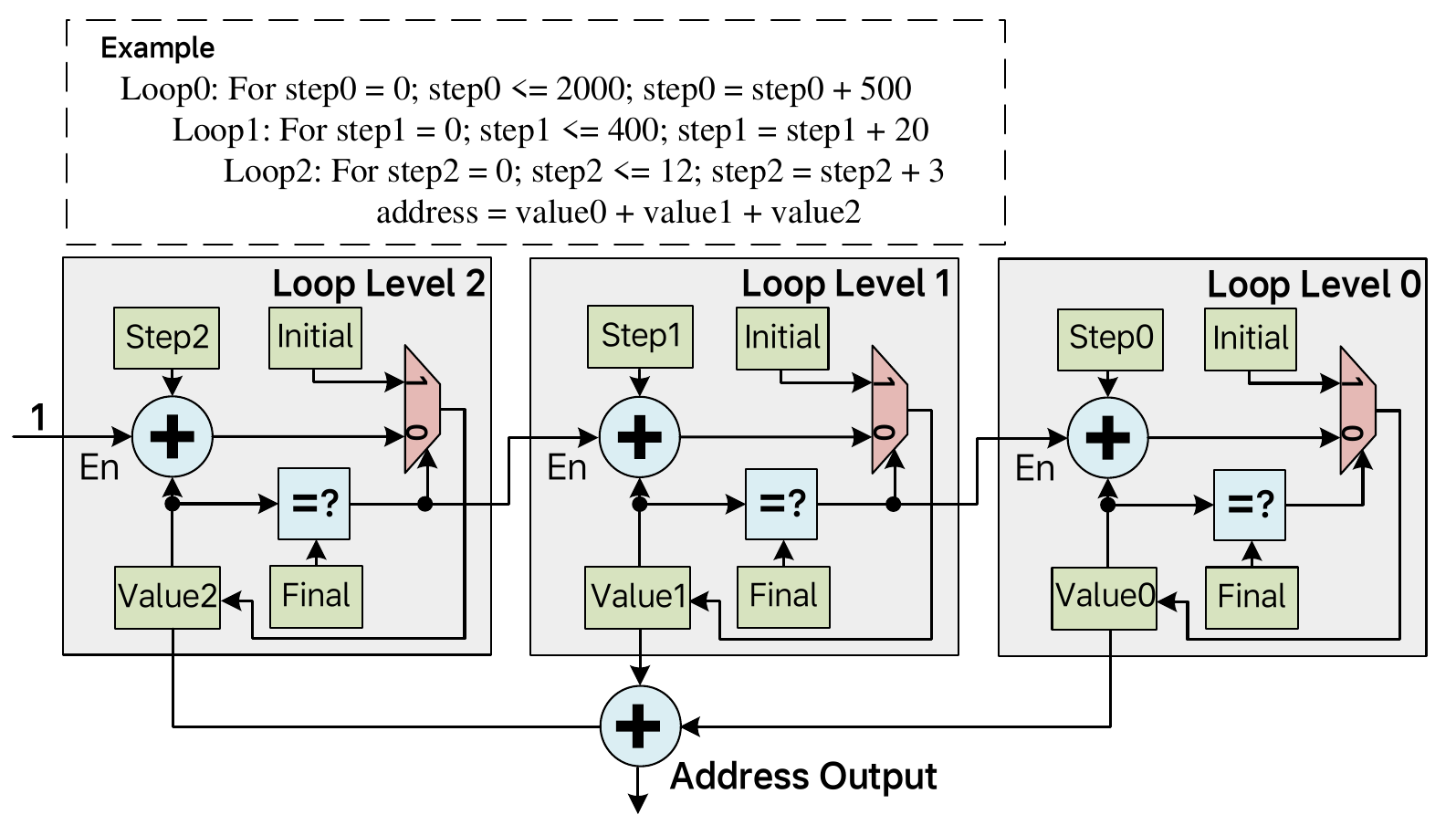}}
	\caption{An example of 3-level TWU.}
	\label{TWU}
    \vspace{-1em}
\end{figure}
\textbf{2) Tensor Walker Unit}


TPB functional units access tensors in HBSM by streaming input data in and output data out. This requires generating address sequences tailored to specific computation patterns. For operations like convolution, addresses often follow complex, non-linear patterns defined by nested loops rather than simple linear increments. To support this, Tensor Walker Units (TWUs) are used to flexibly generate the required address sequences, enabling efficient access to both input activations and weights.

Typically, a TPB functional unit has two or more input TWUs, and one output TWU. The TWU can be configured by the TPB instruction for that functional unit, specifying the number of nested loop levels, and the \textit{Initial}, \textit{Step}, and\textit{ Final} values for each loop level. After configuration, the TWU generates one address at every iteration, until the \textit{Value} counter at every loop level reaches the \textit{Final} at that level. In Fig. \ref{TWU}, an example of 3-level TWU is shown. The output address is the sum of \textit{Value} counters of all loop levels. The \textit{Value }counter of one loop level will be triggered to increment by the \textit{Step }of that loop level, when the \textit{Value }of the inner loop level reaches the \textit{Final}. Of course, the \textit{Value} counter of the inner-most loop increments unconditionally every iteration. Whenever the \textit{Value }counter reaches the \textit{Final }at a certain loop level, it restarts from the \textit{Initial }when it is incremented the next time. \label{Section IV-C-2}The TWU also supports address generation for double buffering. By specifying a \textit{Step} value with a buffer offset at an outer loop level, the programmer can seamlessly alternate between two buffer regions.

TWU's ability to generate rich address patterns—combined with HBSM-based data sharing—enables efficient implementation of complex data communication between TPB functional units without requiring dedicated datapaths or buffers. As such, the TWU is a key enabler of the M100 NPU’s simple yet powerful dataflow architecture.

\textbf{3) Synchronization Unit}


Synchronization is a crucial component of dataflow parallel computing. Functional units must notify peers when data is produced or consumed to maintain flow across pipeline stages. Traditional architectures rely on atomic operations or exclusive load/store instructions for synchronization, which can be inefficient and tightly coupled to cache and memory subsystems. In contrast, synchronization in the M100 NPU—an AI accelerator optimized for dataflow execution—can be greatly simplified and is handled as follows:

\begin{itemize}
\item  One agent updates its own execution state while performing a certain task.
\item The other agent monitors the first agent's state, and decides if it is OK to proceed to the next step.
\end{itemize}


This update/monitor relationship can operate bidirectionally between two agents. For example, a producer updates its data production state while monitoring the consumer's consumption state, and simultaneously, the consumer updates its consumption state while monitoring the producer’s production state. This enables the two to work in tandem as a computation pipeline. The same mechanism can be extended to multiple agents, forming a synchronization network using simple state update and monitor operations.


Within each TPB, the Synchronization Unit (SU) manages hardware counters that track and coordinate execution states. A functional unit claims a counter to update its own progress and can monitor other counters to determine if dependencies are met. Update and monitor actions are triggered at specific execution stages as defined by the TPB instruction. When an update request is issued, the SU increments the assigned counter by one. A monitor request includes an expected value, and the SU responds only if the counter meets or exceeds that value. Until then, the requesting unit pauses execution. Software controls which counters are updated or monitored, and with proper assignment, a highly efficient, synchronized execution pipeline can be achieved across parallel functional units.

\textbf{4) Tensor Computing Unit}

\begin{figure}[ht]
	\centering
	\centerline{\includegraphics[scale=0.38]{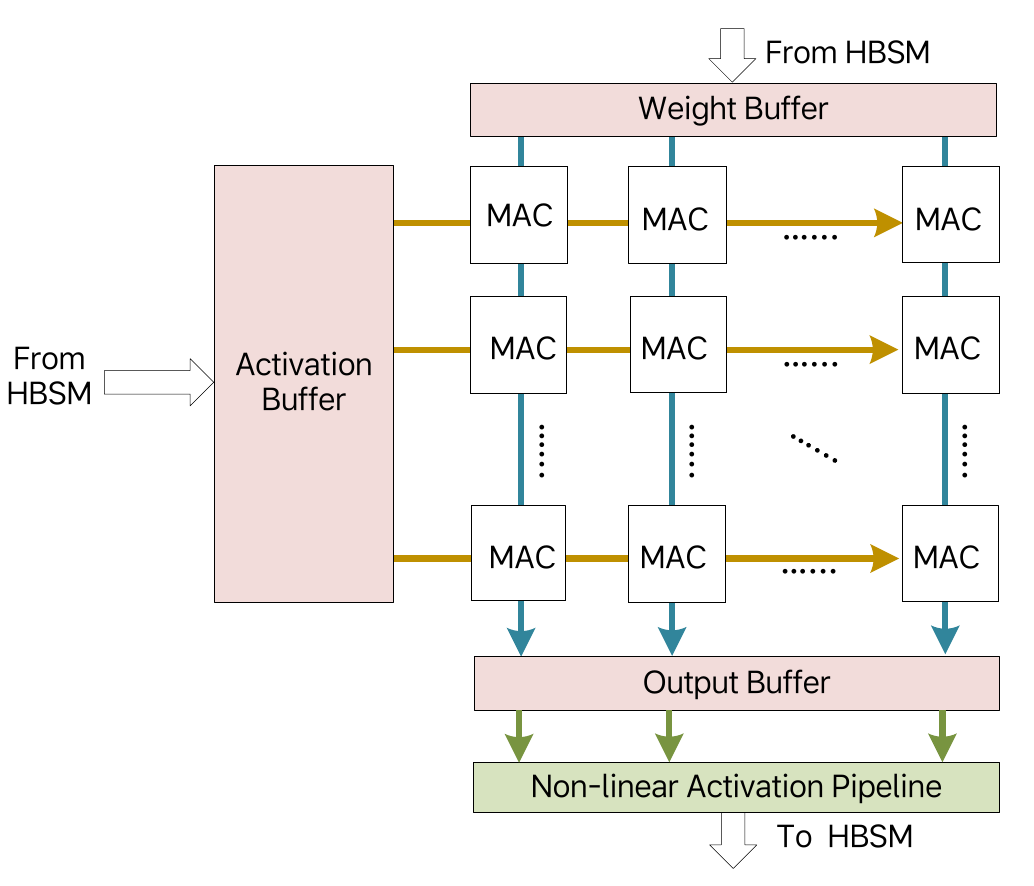}}
	\caption{Architecture of the TCU.}
	\label{TCU}
\end{figure}


The Tensor Computing Unit (TCU) accelerates tensor contraction operations using a dense array of compute elements. To sustain high throughput under limited memory bandwidth, data reuse is essential. As shown in Fig. \ref{TCU}, the TCU arranges Multiply–Accumulate (MAC) units in an 8×64 array. Each MAC performs a 4-element dot product per cycle. Activation data is reused across rows, and weight data across columns. For a matrix multiplication of size 32×32 × 32×64, the computation completes in 32 cycles—matching the 32B/cycle and 64B/cycle input bandwidths of the activation and weight buffers, assuming 1-byte elements. With double buffering, the TCU can sustain peak throughput for both matrix multiplication and convolution operations.


Post-MAC, partial sums are stored in the output buffer and passed through a non-linear activation pipeline before being written to HBSM. Since tensor contractions typically reduce along large axes, output data is smaller and write bandwidth is rarely a bottleneck. For large tensors, TCU includes outer loop control logic to iterate through blocks efficiently, keeping the pipeline active with minimal idle cycles.

\textbf{5) Configurable Vector Unit}

\begin{figure}[ht]
	\centering
	\centerline{\includegraphics[scale=0.36]{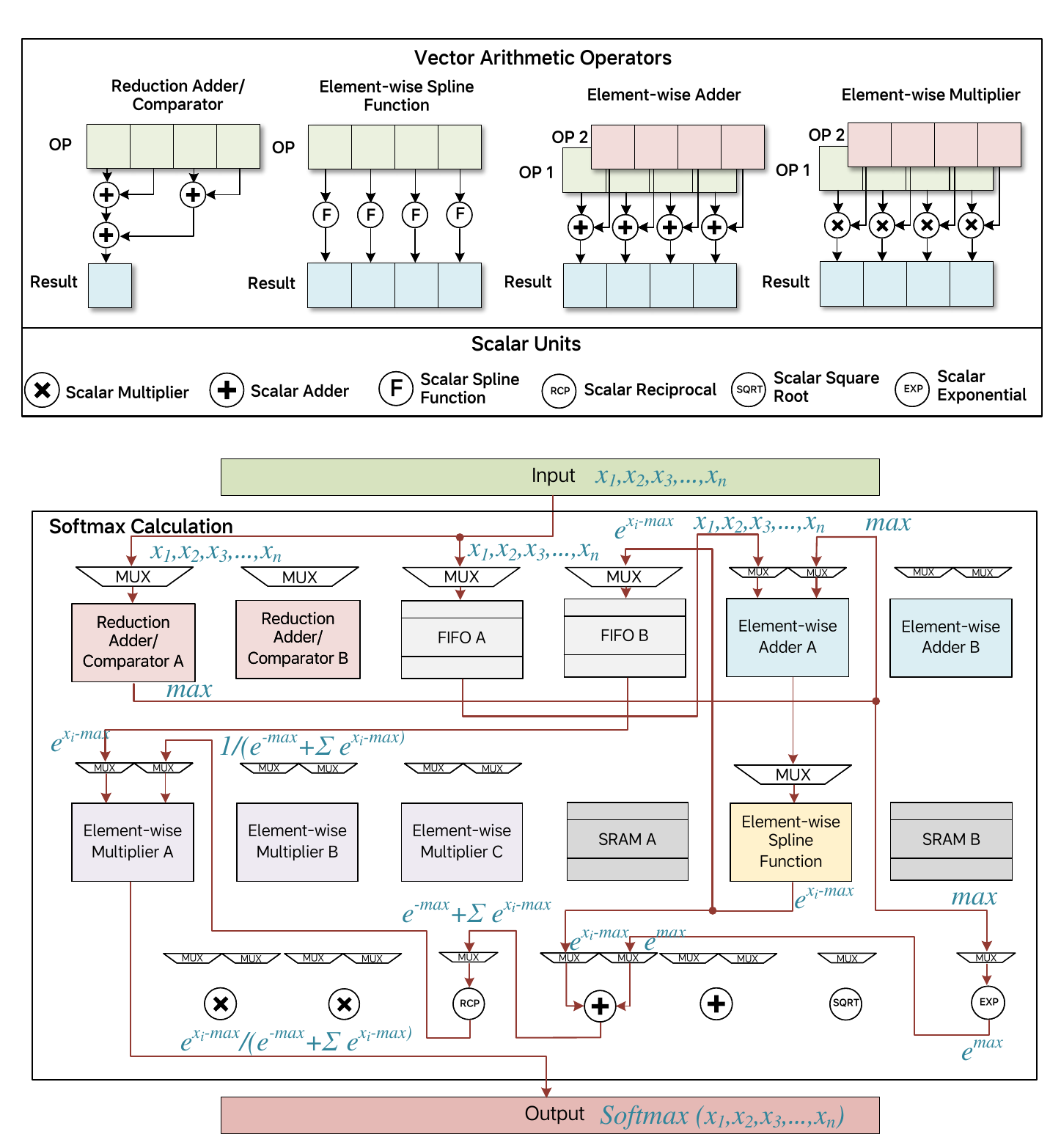}}
	\caption{The architecture of CVU.}
	\label{CVU}
\end{figure}


Figure \ref{CVU} illustrates the core components of the Configurable Vector Unit (CVU), which consists of single-function vector arithmetic operators. Each takes one or two input vector streams and produces a single output stream. TPB instructions can configure the CVU to route inputs through a single operator or build multi-stage pipelines with intermediate buffers. This flexibility enables highly efficient execution of common vector operations. At the bottom of Fig. \ref{CVU}, a CVU configuration is shown for Softmax Computation—a frequently used operator in transformer-based models. \label{ Section IV-C-5}The computation steps across the configured pipeline stages are annotated at the participating vector operators to illustrate how the CVU performs the softmax operation with optimal efficiency.


For more complex vector operations that can't be fully pipelined, the CVU can process them in multiple stages, each handled by a separate TPB instruction. While this may reduce throughput, performance remains comparable to or better than traditional vector cores. Thanks to its vast configuration space, the CVU is well-suited to adapt to diverse vector computation patterns in AI inference workloads.

\textbf{6) DMAs}

In addition to its compute units, each TPB is equipped with high-performance DMA engines to support dataflow both within the TPB and across multiple TPBs. There are two types of DMAs in the TPB:

\begin{itemize}

\item The Data Transformation DMA Unit (DTDU) functions like a compute unit and executes TPB instructions. It handles data movement within HBSM, supports operations like matrix transposition, and can efficiently initialize memory by filling specified address ranges with predefined values.


\item The Gather-Scatter DMA Unit (GSDU) is managed by the cluster CPU and does not execute TPB instructions directly. It handles irregular data movement patterns that are difficult to encode in standard TPB instructions. Instead, a TPB instruction triggers the CPU Starter Unit (CSU), which launches a CPU routine to control the GSDU. The GSDU transfers data between the local HBSM and external memories—such as another TPB’s HBSM, CCB SRAM, or DDR. As its name suggests, it supports gather and scatter operations between local and remote memory spaces.

\end{itemize}

\textbf{7) CPU Starter Unit }


The CSU executes a TPB instruction that triggers an interrupt to request assistance from the cluster CPU. Task parameters are stored in the CSU, and the CPU's interrupt service routine retrieves them to determine the required operation. Tasks may involve scalar or vector processing, or control of the GSDU via the VCIX interface for runtime-determined data movement. Once the routine completes, it notifies the CSU, which then marks the TPB instruction as finished.

\section{Compiler and Runtime Software Stack}
\label{SoftwareStack}
As part of a vertically integrated solution, the M100 compiler and runtime software stack play a crucial role in ensuring correct functionality and excellent performance. 

\subsection{Compiler}
As shown in Fig. \ref{Compilation architecture}, the M100 AI compiler toolchain includes a space-time scheduler, a graph compiler, and a backend compiler:
\begin{figure}[ht]
	\centering
	\centerline{\includegraphics[scale=0.46]{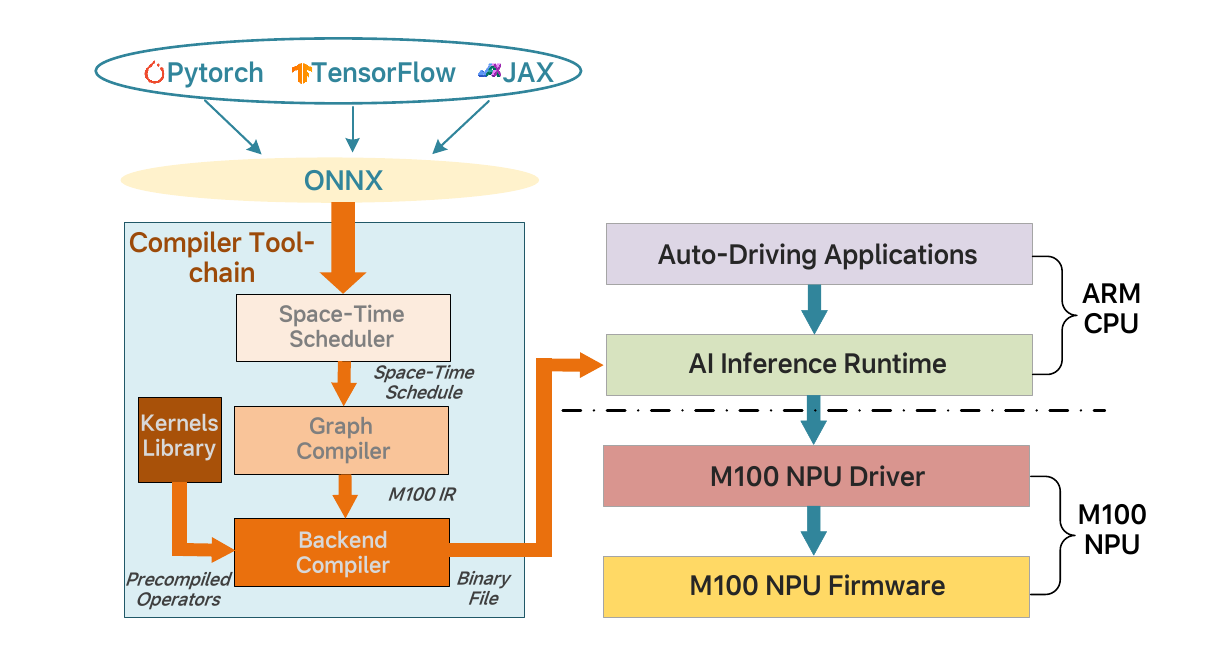}}
	\caption{Overview of  M100 AI compiler toolchain.}
	\label{Compilation architecture}
    \vspace{-1em}
\end{figure}

\begin{figure}[ht]
   	\centering
   	\centerline{\includegraphics[scale=0.25]{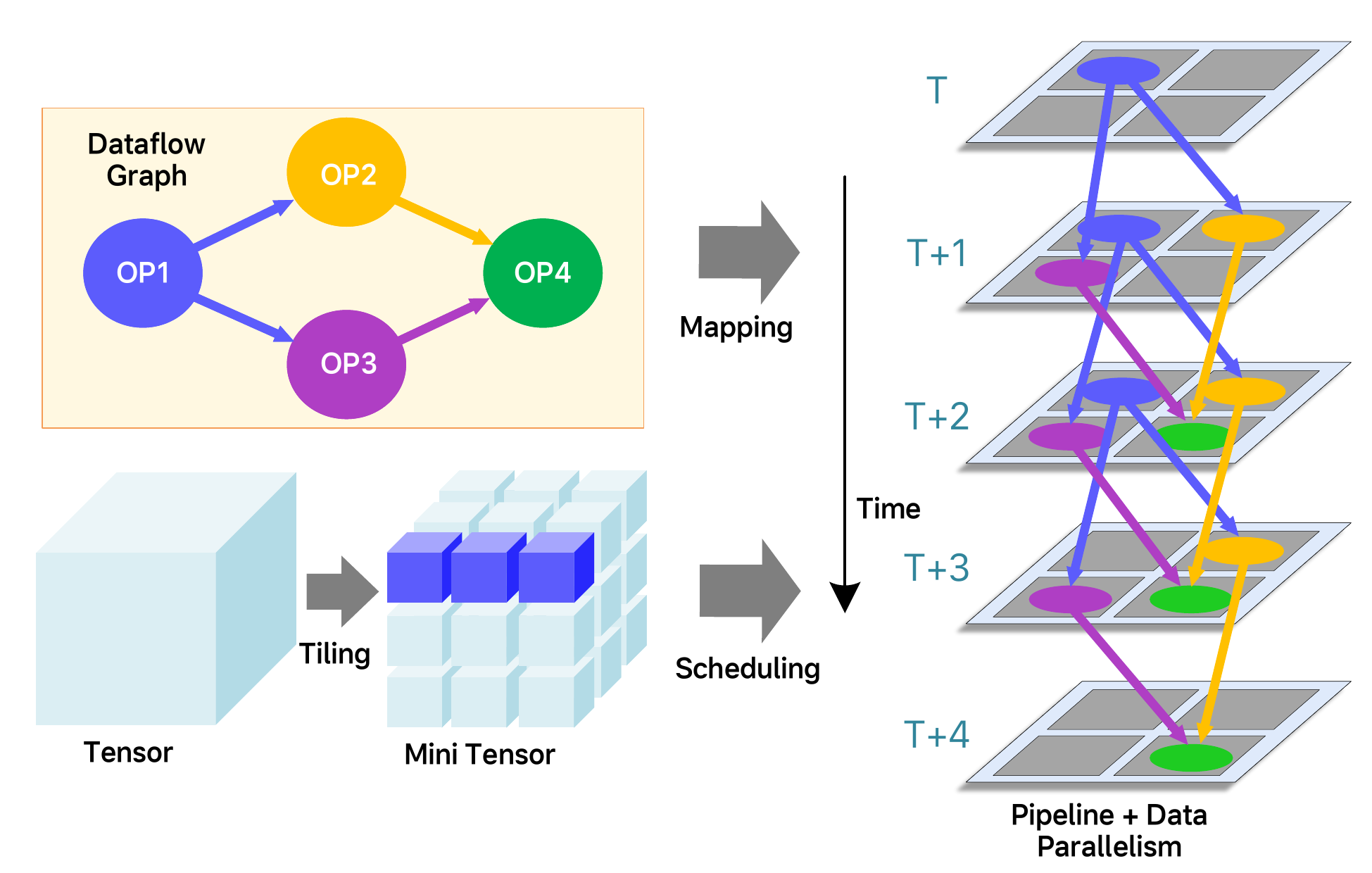}}
   	\caption{ Space-time scheduler subgraph mapping and tensor streaming on M100.}
   	\label{Space-time Compile}
   \end{figure}

\begin{itemize}
	\item The space-time scheduler maps a neural network subgraph onto the M100 NPU hardware. If necessary, large tensors are partitioned into mini-tensors that are passed along the processing pipeline constructed by the space-time scheduler of the dataflow compiler. An example of space-time scheduling is illustrated in Fig. \ref{Space-time Compile}. A subgraph containing four computational operators (OP1, OP2, OP3, OP4) is distributed into four TPBs spatially. The input tensor undergoes dimensional decomposition along multiple axes, producing a number of mini-tensors that are subsequently streamed through the allocated TPBs following temporally scheduled phases.
    
	\item The graph compiler performs graph optimization and dynamic memory allocation for dynamic tensors. The graph optimization includes tensor fusion, dead code elimination, algebraic simplification, layout transformation, and so on.
	
    \item The back-end compiler is a C-extended compiler that generates intrinsic instructions that utilize hardware capabilities of the M100 architecture, such as tensor computing, data movement, and synchronization.
\end{itemize}

\subsection{Runtime Software}
The M100 runtime software stack includes AI inference runtime and NPU driver running on ARM Cortex-A78 cores of the SoC, and NPU firmware running on NPU RISC-V cores. The AI inference runtime is responsible for preparing input data, loading NN models, kicking off tasks with allocated resources, and post-processing inference results for downstream applications. The AI inference runtime also monitors any error or exception state that the NPU runs into, and makes sure the NPU meets the automotive functional safety (FuSa) requirements. The NPU driver serves as the hardware abstraction layer for high-level application software. The NPU firmware employs just-in-time (JIT) compilation techniques, dynamically generating optimized TPB instructions based on the binary code generated by the M100 compiler toolchain. The firmware also performs on-the-fly computation of tensor shapes and memory addresses for tensor storage. The TPB instructions are issued by the NPU firmware to a group of TPBs assigned to a task.

\section{Evaluation Results}
\label{EvaluationResults}

To evaluate the performance of the M100 NPU architecture, we conducted a comparative study between M100 and NVIDIA Thor-U, an advanced SoC platform developed for AD and edge AI inference, using benchmarks relevant to AD applications. In this section, we first introduce the hardware configurations of both platforms. Then, the characteristics of the selected benchmarks are analyzed. Finally, performance data is presented to demonstrate how the M100 achieves competitive—or superior—AI inference efficiency and hardware utilization in key AD workloads compared to Thor-U.

\subsection{Hardware Configurations}

At the time of writing, Li Auto has not formally disclosed detailed performance specifications for the M100 beyond basic metrics such as DDR bandwidth and die size. Table \ref{hardaware configuration comparison} presents the available data for the M100 alongside comparable figures for Thor-U. Both platforms offer identical DDR bandwidth, while Thor-U features a slightly larger die size—suggesting comparable raw computing capabilities. To ensure fairness, performance data was collected from both platforms using an identical power budget during benchmarking.

\renewcommand{\arraystretch}{1.2}

\begin{table}[ht]
	\centering
	\caption{Comparison of Hardware Configurations between NVIDIA Thor-U and M100}
	\begin{tabular}{|l|ll|}
		\hline
		Metric                                 & \multicolumn{1}{l|}{Thor-U}            & M100               \\ \hline
		\begin{tabular}[c]{@{}l@{}}
        DDR Memory Bandwidth\end{tabular}      & \multicolumn{1}{l|}{273 GB/s}          & 273 GB/s           \\ \hline

        Die Size                               & \multicolumn{1}{l|}{415 $mm^2$}           & 399.8 $mm^2$                      \\ \hline
		
		Process                                & \multicolumn{1}{l|}{TSMC N4}           & TSMC N5A           \\ \hline
	\end{tabular}
	\label{hardaware configuration comparison}
\end{table}

\subsection{Benchmarks}

Autonomous driving and intelligent cockpit are two important features of modern-day smart vehicles. We selected UniAD\cite{hu2023planning}, the state‑of‑the‑art end‑to‑end AD algorithm, as our autonomous driving benchmark application. For intelligent cockpit scenarios, LLMs like LLaMA2‑7B\cite{touvron2023llama} are crucial components to support intelligent interactions between the vehicle and the driver/passenger. Therefore, we picked LLaMA2‑7B as another important benchmark for performance evaluation. 
Additionally, to comprehensively evaluate the performance of integrated VLA capabilities in AD scenarios, we selected a critical component of Li Auto’s in-house developed MindVLA model as the third benchmark.

While porting UniAD, LLaMA2‑7B, and MindVLA to the Thor-U and M100 platforms, we ensured that comparable computing resources and power consumption were maintained during benchmark execution on both systems. Further details of these three benchmarks are provided in the remainder of this section.

\subsubsection{Model Architecture}

\begin{itemize}

\item To better represent the currently deployed AD algorithm by Li Auto, the UniAD benchmark was modified by replacing ResNet‑101 with RegNet. As illustrated in Fig. \ref{UniAD Pipeline}, the UniAD algorithm provides a unified framework that seamlessly integrates two core tasks of autonomous driving: perception and prediction. Perception covers object detection and tracking, and prediction handles motion forecast and occupancy prediction. Both perception (BevFormer, TrackFormer, MapFormer) and prediction (MotionFormer, OccFormer) modules are based on transformer architectures. These components are connected through a large number of query tokens (e.g., 900 in TrackFormer), enabling ample opportunities for parallel processing during inference. Before reaching the first perception stage (BEVFormer), a CNN-based backbone is used to extract features from the input images.

 \begin{figure*}[ht]
	\centering
	\centerline{\includegraphics[scale=0.21]{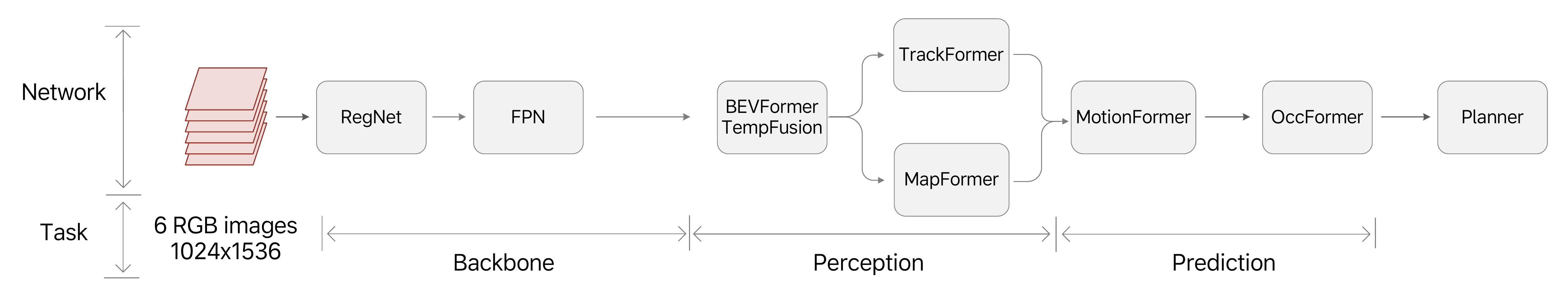}}
	\caption{ UniAD Framework.}
	\label{UniAD Pipeline}
    \vspace{-1em}
\end{figure*}

\item LLaMA2‑7B is a transformer-based large language model with approximately 7 billion parameters. It employs a standard decoder-only transformer architecture featuring multi-head self-attention and feed-forward networks. The inference process consists of two stages: a prefill phase, which processes the input sequence in parallel, and a decode phase, which generates tokens sequentially. 

\item MindVLA, Li Auto's next generation autonomous driving algorithm, integrates LLM components with a Mixture-of-Experts (MoE) transformer architecture to improve both model capacity and inference efficiency.
\end{itemize}

\subsubsection{Computation Complexity Analysis}

TABLE \ref{UniAD Parameters} provides an overview of the parameter counts and MAC operations associated with each network model in UniAD. The CNN-based backbone accounts for most of computational resource usage, primarily due to intensive convolution operations on high-resolution images.  In real-world driving scenarios, perception tasks (BEVFormer, TrackFormer, MapFormer) typically operate at higher frame rates than prediction tasks (MotionFormer, and OccFormer) and, consequently, demand greater computational resources.  Therefore, our analysis focuses on the CNN-based backbone and the transformer-based perception models within UniAD.

\renewcommand{\arraystretch}{1.2}
\begin{table}[ht]
	\centering
	\caption{Parameter Sizes and MAC Counts of Network Models in UniAD}
	\begin{tabular}{|lll|l|l|}
		\hline
		\multicolumn{1}{|l|}{Module}                      & \multicolumn{1}{l|}{\begin{tabular}[c]{@{}l@{}}Network\\ Architecture\end{tabular}}               & \begin{tabular}[c]{@{}l@{}}Network\\ Model\end{tabular} & \begin{tabular}[c]{@{}l@{}}Parameters\\ (M)\end{tabular} & \begin{tabular}[c]{@{}l@{}}MAC \\ Operations\\ (GFLOPS)\end{tabular} \\ \hline
		\multicolumn{1}{|l|}{\multirow{3}{*}{Backbone}}   & \multicolumn{1}{l|}{CNN Based}                                                                    & RegNet + FPN                                            & 30                                                       & 2381.6                                                               \\ \cline{2-5} 
		\multicolumn{1}{|l|}{}                            & \multicolumn{1}{l|}{\multirow{6}{*}{\begin{tabular}[c]{@{}l@{}}Transformer\\ Based\end{tabular}}} & BEVFormer                                               & 85.6                                                     & 1492.9                                                               \\ \cline{3-5} 
		\multicolumn{1}{|l|}{}                            & \multicolumn{1}{l|}{}                                                                             & TempFusion                                              & 0.2                                                      & 49.0                                                                 \\ \cline{1-1} \cline{3-5} 
		\multicolumn{1}{|l|}{\multirow{2}{*}{Perception}} & \multicolumn{1}{l|}{}                                                                             & TrackFormer                                             & 8.5                                                      & 97.17                                                                \\ \cline{3-5} 
		\multicolumn{1}{|l|}{}                            & \multicolumn{1}{l|}{}                                                                             & MapFormer                                               & 6                                                        & 105.94                                                               \\ \cline{1-1} \cline{3-5} 
		\multicolumn{1}{|l|}{\multirow{2}{*}{Prediction}} & \multicolumn{1}{l|}{}                                                                             & MotionFormer                                            & 22.6                                                     & 266.55                                                               \\ \cline{3-5} 
		\multicolumn{1}{|l|}{}                            & \multicolumn{1}{l|}{}                                                                             & OccFormer                                               & 46.2                                                     & 687.62                                                               \\ \hline
		\multicolumn{3}{|l|}{Planner}                                                                                                                                                                                   & 3.5                                                      & 220.75                                                               \\ \hline
	\end{tabular}
	\label{UniAD Parameters}
\end{table}

LLaMA inference consists of two phases: prefill and decode. In the prefill phase, all tokens in the input sequence are processed in parallel, and the large number of concurrent tokens—similar to the parallel queries in UniAD—offers a high degree of computational parallelism. In contrast, the decode phase generates one token per step, resulting in limited parallelism and making it a memory-bound operation.

Unlike LLaMA2‑7B, the LLM component of MindVLA employs a Mixture‑of‑Experts (MoE) strategy with 8 experts. For evaluation, we use a 431‑million‑parameter configuration as our benchmark.



\subsection{Experimental Results}

It is worth noting that our experiments utilized 12 out of the 14 available clusters on the M100 NPU, representing 86\% of its total computing capacity. This configuration is designed to ensure higher chip yield by allowing up to two defective clusters. For chips with fewer than two defective clusters, even greater performance can be achieved by leveraging the additional hardware resources.

\subsubsection{UniAD}

TABLE \ref{UniAD Comparison} compares the results of six UniAD benchmarks running on the M100 and Thor-U platforms. 
For the M100 platform, we utilized 8 out of the 14 available compute clusters for UniAD tasks, while reserving the remaining 6 clusters for other cockpit domain functions. This allocation strategy demonstrates the M100's capability to handle multiple domain-specific workloads simultaneously while maintaining performance isolation.

The results show that M100 achieves speedups ranging from 1.2× to 6.3× across different network components, with most modules showing 3.8× to 4.4× performance. Even with only 8 clusters dedicated to AD tasks, the M100 sustains 30 FPS for perception tasks, meeting the real-time requirements of autonomous driving. In contrast, Thor-U delivers only 7.9 FPS, which falls short of the performance needed to deploy Navigate on Autopilot in high-speed driving scenarios.

Under the same power budget, the M100 delivers a 3.8× higher frame rate than the Thor-U. This performance gain is attributed to M100’s tightly integrated hardware–software solution for parallelizing AI inference tasks. Specifically, its compiler-generated, carefully orchestrated dataflow execution enables an exceptionally high degree of parallelism across computation and data movement units, while incurring minimal synchronization overhead.

\begin{table}[ht]
    \centering
	\caption{Performance Comparison of Different Networks in UniAD on M100 and Thor-U}
	\begin{tabular}{|l|l|l|l|}
		\hline
		& \begin{tabular}[c]{@{}l@{}}M100\\ (with 8 clusters active)\end{tabular} & Thor-U   & \begin{tabular}[c]{@{}l@{}}M100\\ speedup\end{tabular} \\ \hline 
		RegNet      & 13.1 ms                                                                 & 57.4 ms  & 4.4x                                                   \\ \hline
		FPN         & 4.23 ms                                                                 & 5.1 ms   & 1.2x                                                   \\ \hline
		BEVFormer   & 7.92 ms                                                                 & 32.83 ms & 4.1x                                                   \\ \hline
		TempFusion  & 4.47 ms                                                                 & 17 ms    & 3.8x                                                   \\ \hline
		TrackFormer & 1.27 ms                                                                 & 7.95 ms  & 6.3x                                                   \\ \hline
		MapFormer   & 1.46 ms                                                                 & 6.14 ms  & 4.2x                                                   \\ \hline
		Frame rate  & 30 FPS                                                                  & 7.9 FPS  & 3.8x                                                   \\ \hline
	\end{tabular}
	\label{UniAD Comparison}
\end{table}

\begin{figure}[ht]
	\centering
	\centerline{\includegraphics[scale=0.3]{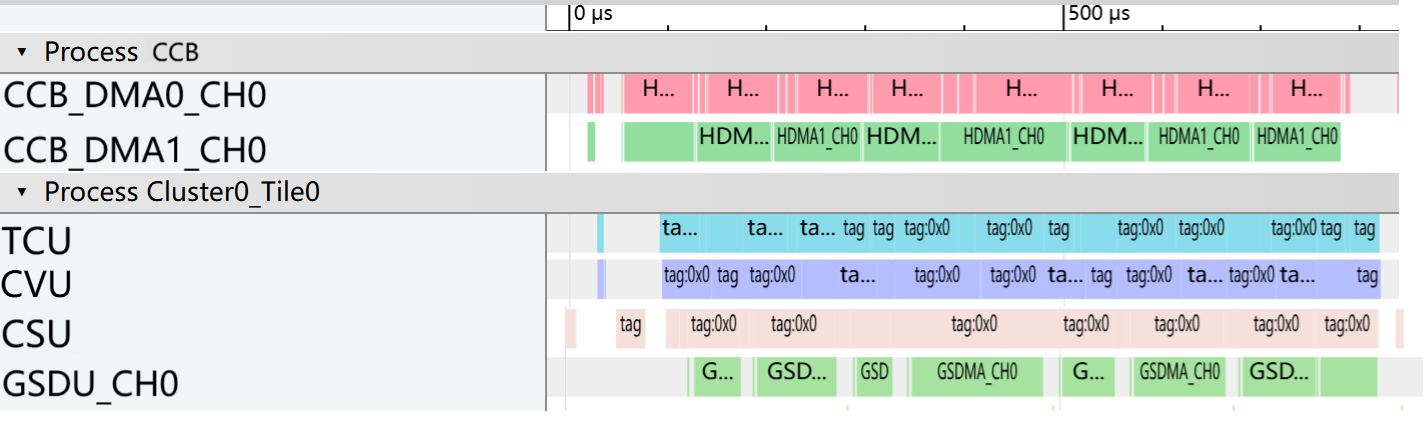}}
	\caption{ Detailed execution trace of M100 TPB instructions collected by the in-house profiling tool.}
	\label{Trace}
\end{figure}

We collected detailed execution timeline data by using our in-house performance profiling software to trace the M100’s behavior. The resulting execution timeline is shown in Fig. \ref{Trace}.  Different colored blocks represent the activity of various processing units over time: continuous segments indicate sustained activity, while gaps denote idle or waiting periods. In this trace, throughout most of the sampling window, the DMAs in the CCB—along with the TCU, CVU, CSU, and GSDU in one of the TPBs—remain continuously active, with substantial overlap in task execution. This indicates high hardware utilization and highlights the M100 architecture’s strong parallel execution capabilities and overall efficiency.

\subsubsection{LLaMA2-7B}

In the LLaMA2‑7B benchmark setup, the input sequence length is set to 1,024 tokens. TABLE \ref{LLaMA comparison} summarizes the performance comparison between the M100 and Thor-U platforms for both the decode and prefill phases of the inference task. For the decode phase, we employ W4A16 quantization, where weights are represented as 4-bit integers and activation features as 16-bit floating-point values. The M100 achieves a latency of 21.34ms, demonstrating performance comparable to Thor-U’s 20ms. Although Thor-U holds a slight advantage in this metric, it is largely due to the extensive optimizations developed for open-source models like LLaMA2‑7B on NVIDIA platforms. On the other hand, since the M100 and Thor-U platforms share identical DDR memory bandwidth—which primarily bounds the performance of the decode stage—comparable performance between the two platforms is expected. For the prefill phase, we apply W8A8 quantization, representing both weights and activations as 8-bit integers. The M100 demonstrates a significant advantage, completing inference in 79ms compared to 154ms on Thor-U—achieving a 1.95× speedup. This improvement is attributed to the M100’s highly efficient tensor processing units and the dataflow-driven synchronization mechanism that enables seamless coordination among them.


\begin{table}[ht]
	\centering
    \caption{Performance Comparison of LLaMA2-7B Inference Phases on M100 and Thor-U}
	\begin{tabular}{|l|l|l|l|}
		\hline
		& \begin{tabular}[c]{@{}l@{}}M100\\ ( with 12 clusters active)\end{tabular} & Thor-U        & \begin{tabular}[c]{@{}l@{}}M100 \\ speedup\end{tabular} \\ \hline
		decode  & 21.34 ms (W4A16)                                                          & 20 ms (W4A16) & 0.94x                                                   \\ \hline
		prefill & 79 ms (W8A8)                                                              & 154 ms (W8A8) & 1.95x                                                   \\ \hline
	\end{tabular}
	\label{LLaMA comparison}
\end{table}




\subsubsection{MindVLA (LLM Part)}

In addition to evaluating open-source models, we also tested MindVLA, Li Auto’s next-generation autonomous driving model developed in-house. This evaluation demonstrates the M100 platform’s capability to support production-level AD applications. Table \ref{MindVLA} presents a performance comparison between the M100 and Thor-U platforms for the LLM component of MindVLA.
\begin{table}[ht]
	\centering
	\caption{Performance Comparison of MindVLA (LLM Component) on M100 and Thor-U}
	\begin{tabular}{|l|l|l|l|}
		\hline
		& \begin{tabular}[c]{@{}l@{}}M100\\ ( with 12 clusters active)\end{tabular} & Thor-U  & \begin{tabular}[c]{@{}l@{}}M100 \\ speedup\end{tabular} \\ \hline
		decode  & 0.1 ms                                                                    & 0.3 ms  & 3x                                                      \\ \hline
		prefill & 0.84 ms                                                                   & 1.74 ms & 2.1x                                                    \\ \hline
	\end{tabular}
	\label{MindVLA}
\end{table}

For the decode phase, the M100 achieves a latency of 0.1ms, compared to 0.3ms on the Thor-U, resulting in a 3× speedup. In the prefill phase, the M100 completes inference in 0.84ms versus 1.74ms on the Thor-U, achieving a 2.1× speedup. While only the performance of the LLM component is presented here, these results highlight the M100’s advantages in supporting more advanced autonomous driving workloads.


\section {Conclusion}
\label{Conclusion}

We presented the M100 SoC and NPU—Li Auto's solution for addressing general-purpose AI inference workloads—built on a dataflow architecture that reduces design complexity by enabling the compiler and runtime software to orchestrate computation and data movement across processing elements. We detailed the architecture of key functional blocks and explained the rationale behind major design decisions. Comparative evaluation results demonstrate that the M100 NPU outperforms leading GPGPU platforms by a significant margin, without compromising flexibility. We believe that by striking an effective balance between software and hardware design complexity, the classic dataflow architecture can be revitalized to meet the rapidly evolving demands of modern AI computing.


\bibliographystyle{IEEEtran}
\bibliography{refs}

\end{document}